\documentclass[11pt]{article}

\usepackage[T1]{fontenc}
\usepackage[utf8]{inputenc}
\usepackage[letterpaper,margin=1in]{geometry}
\usepackage{amsmath}
\usepackage{amssymb}
\usepackage{amsthm}
\usepackage{array}
\usepackage{bm}
\usepackage{booktabs}
\usepackage{graphicx}
\usepackage{longtable}
\usepackage{microtype}
\usepackage[round,authoryear]{natbib}
\usepackage{placeins}
\usepackage{fvextra}
\usepackage{xcolor}
\usepackage{hyperref}

\graphicspath{{figures/}}

\hypersetup{
    colorlinks=true,
    citecolor=blue,
    linkcolor=blue,
    urlcolor=blue
}

\title{Hypothesize, Evaluate, Refine: A Scientific Agent for PDE Discovery
with Unknown Spatial Coefficient Fields}
\author{
    YuJie Huang \\
    Fujian University of Technology \\
    \texttt{YuJieHuang0526@gmail.com}
    \and
    WenWu He \\
    Fujian University of Technology  \\
    Fujian Provincial Key Laboratory of Big Data Mining and Applications  \\
    \texttt{hwwhbb@163.com}
    \and
    ZhuoEr Lin \\
    Fujian University of Technology \\
    \texttt{zel755158751@gmail.com}
    \and
    Congcong Liu \\
    Shenzhen Institutes of Advanced Technology, Chinese Academy of Sciences \\
    \texttt{cc.liu@siat.ac.cn}
    \and
    Dong Liang \\
    Shenzhen Institutes of Advanced Technology, Chinese Academy of Sciences \\
    \texttt{dong.liang@siat.ac.cn}
    \and
    Zhuo-Xu Cui \\
    Shenzhen Institutes of Advanced Technology, Chinese Academy of Sciences \\
    \texttt{zx.cui@siat.ac.cn}
}

\date{August 2026}

\begin{document}

\maketitle

\begin{abstract}
Discovering PDEs in heterogeneous media requires jointly identifying the
governing operator and the unknown spatial fields that parameterize it. These
tasks are coupled: changing field placement changes the differential law,
while a sufficiently flexible field can conceal structural error on a single
trajectory. We present Hypothesize, Evaluate, Refine for PDE Discovery
(HER-PDE), a scientific-agent framework that discovers compositional PDE
structure together with nonparametric, time-invariant coefficient fields. The
Agent analyzes two noisy trajectories generated by different excitations,
proposes complete expression-tree hypotheses, and combines creative structural
exploration with local candidate refinement. Its Hypothesis Evaluation
Interface (HEI) estimates only the fields explicitly declared in each
hypothesis, never adds missing terms, and scores structures by bidirectional
cross-excitation transfer. The selected law is subsequently audited on a
sealed temporal interval. Across five controlled two-dimensional systems
observed with $5\%$ relative Gaussian state noise, the Agent recovers the
generating operator in all five cases, including equivalent signed-field and
product-rule parameterizations. Across nine unknown coefficient fields, the
recovered fields attain a median Pearson correlation of approximately $0.85$
and a median relative $L_2$ error of approximately $0.28$. These results show
that agent-guided hypothesis refinement can recover heterogeneous governing
laws without prescribing a parametric form for their spatial coefficients.
\end{abstract}

\section{Introduction}
\label{sec:introduction}

Partial differential equations provide compact models of transport, diffusion,
reaction, and other processes that evolve across space and time. Data-driven
equation discovery seeks these laws directly from observed trajectories, aiming
to recover symbolic models that can be inspected and tested against independent
observations \citep{schmidt_lipson_2009,sindy_2016,pde_find_2017}. In a
heterogeneous medium, the properties that govern the dynamics may themselves
vary across space. Conductivity, diffusivity, transport velocity, and reaction
rate then
enter the law as coefficient fields rather than a few scalar parameters. The
discovery target consequently contains two kinds of unknowns: a discrete
differential operator and the spatial functions that parameterize it.

These two inference problems are coupled. The placement of a field can change
the governing operator: multiplying a Laplacian by $c(\bm{x})$ is different
from placing the same field inside a divergence, where derivatives of $c$ also
enter the law. A candidate may reuse one field across several directions or
assign independent fields to them, and each field may depend on a different
subset of the spatial coordinates. A sufficiently flexible field can also
compensate for an incorrect operator on one observed trajectory. Joint
discovery must therefore recover the operator together with the identity,
placement, dependencies, and values of its fields.

Existing methods often fix either the structural or coefficient representation.
Sparse methods select terms from a supplied library, while methods for variable
coefficients estimate functions within prescribed feature sets
\citep{pde_find_2017,sgtr_2019,luo2023physics,wg_ident_2025}. Open-form symbolic
methods construct expressions beyond a fixed term matrix
\citep{sga_pde_2022,discover_2024,eqgpt_2025}. The underexplored combination is
compositional PDE search in which each candidate declares its coefficient-field
identity, placement, operator scope, spatial dependence, and reuse, followed by
nonparametric field estimation and a test of transfer across independent
excitations. Each candidate structure in this setting defines its own function
estimation problem.

We study this problem using two noisy trajectories generated by different
excitations of the same heterogeneous system. The trajectories have different
state histories but share one governing operator and the same time-invariant
spatial coefficient fields,
$c_k=c_k(\bm{x}_{S_k})$ with
$S_k\subseteq\{1,\ldots,d\}$. Different excitations expose different state and
derivative combinations while leaving the shared fields fixed. A field fitted
on one trajectory can therefore be evaluated by applying the induced law to the
other, in both directions. Bidirectional transfer uses repeated experiments as
structural evidence and tests whether the fitted fields describe the invariant
mechanism shared by both records.

We introduce \emph{Hypothesize, Evaluate, Refine for PDE Discovery}
(\emph{HER-PDE}), a framework built around a scientific Agent
that treats a complete expression tree and its unknown coefficient fields as
one hypothesis. The candidate language preserves derivative order and direction,
nonlinear operations, and candidate-specific operator--field structure. It
admits expressions that are nonlinear in the measured state and its
derivatives while remaining linear in the unknown fields and their induced
spatial derivatives once the tree is fixed. The Hypothesis Evaluation Interface
(HEI) preserves each submitted tree during compilation, fits only the fields
declared by that tree, and measures bidirectional transfer between the two
trajectories. The Agent uses these evaluations to construct competing
hypotheses, revise promising structures, and select a final law among the
evaluated candidates. That law is then audited on a temporal interval kept
sealed throughout search and model selection.

This work contributes:

\begin{itemize}
    \item A joint formulation of PDE structure and unknown, time-invariant
    spatial coefficient fields, together with a language of complete expression
    trees that retains operator scope, field placement, spatial dependencies,
    and field reuse throughout search and evaluation.

    \item HEI for complete hypotheses with unknown fields. It compiles each tree
    into a joint linear problem for the declared fields, includes the analytic
    field derivatives required by operator scope, uses symmetric
    cross-trajectory transfer during search, and audits the selected law on a
    sealed temporal interval.

    \item An evidence-guided two-phase Agent that generates global alternatives,
    revises them using transfer scores and field diagnostics, and selects among
    hypotheses evaluated by HEI.
\end{itemize}

We report five controlled two-dimensional case studies with $5\%$ relative
Gaussian state noise, including four with non-analytic Mat\'ern-$3/2$
coefficient fields. In these studies, the Agent recovers the generating operator
up to field renaming, field-sign reparameterization, or exact product-rule
expansion. The cases span conservative and nonconservative operator placement
as well as shared and independent fields.

\section{Related Work}
\label{sec:related-work}

\begin{figure}[!ht]
    \centering
    \includegraphics[width=0.94\linewidth]{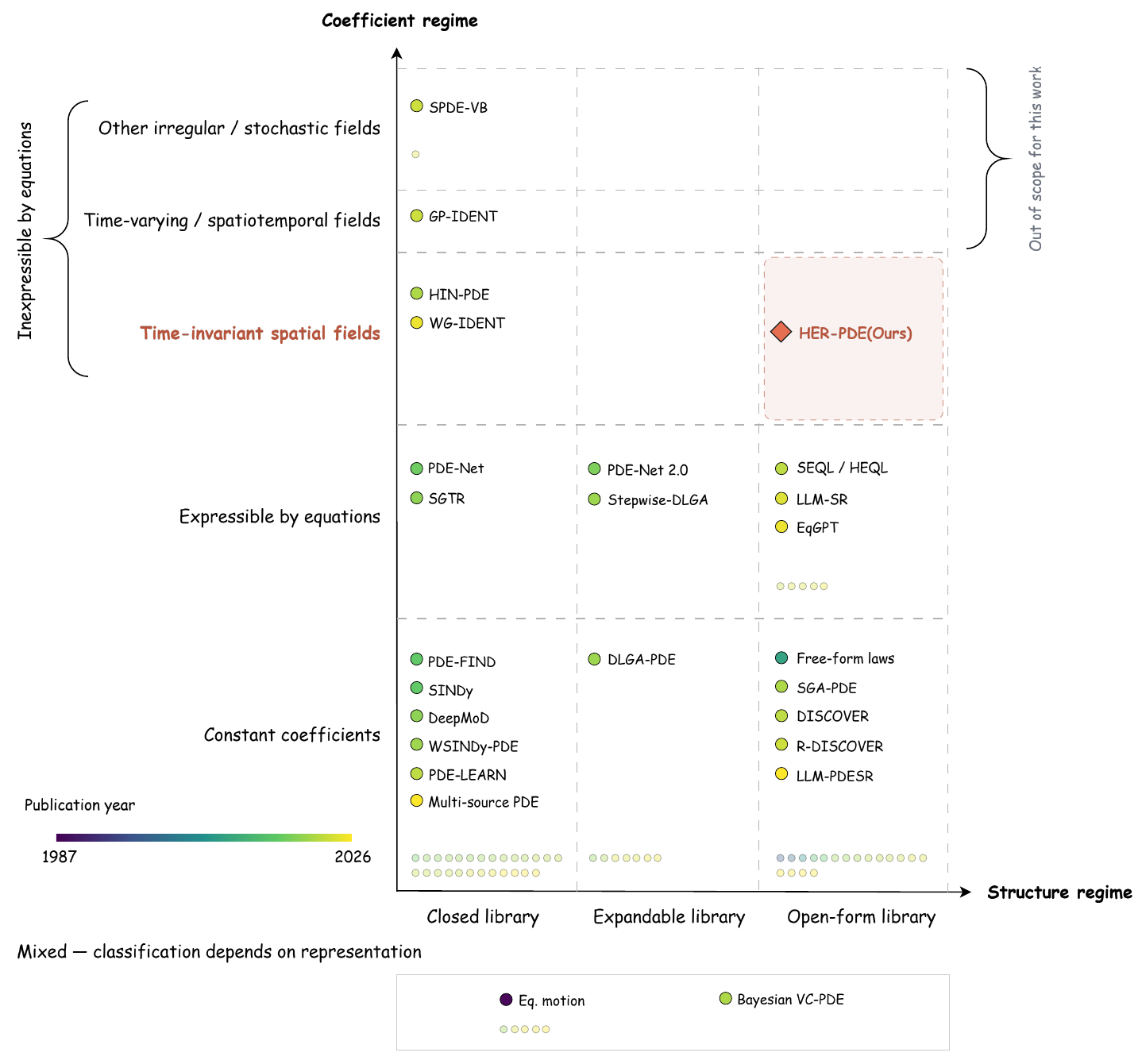}
    \caption{Positioning of equation-discovery methods by structure
and coefficient regime. Labels mark 25 representative works; faint markers
show the remaining 62 publications. Color indicates publication year.
Within-cell placement has no quantitative meaning. HER-PDE targets
open-form PDE discovery with nonparametric, time-invariant spatial coefficient
fields.}
    \label{fig:field-eqgpt-positioning}

\end{figure}

Data-driven equation discovery depends on two coupled representations: how the
governing structure is constructed and how its coefficients are modeled.
Following \citet{lou_review_2026}, we distinguish closed libraries, expandable
libraries, and open-form grammars. We refine the coefficient axis into
constants, compact equation-expressible coefficients, time-invariant spatial
fields, time-varying or spatiotemporal fields, and irregular or stochastic
fields. The final three distinguish variation patterns within Lou et al.'s
``inexpressible by equations'' regime, where the target field is not assumed to
follow a compact explicit law. The subdivision records dependence pattern, not
a quantitative complexity score.
Figure~\ref{fig:field-eqgpt-positioning} summarizes the literature and marks the
regime studied here.

\subsection{Representations of governing structure}
\label{sec:related-work-structure}

\paragraph{Closed libraries.}
Sparse identification casts discovery as term selection from a finite matrix of
candidate functions. SINDy introduced this formulation for dynamical systems,
and PDE-FIND extended it to nonlinear functions and spatial derivatives
\citep{sindy_2016,pde_find_2017}. Later work improved how the library is
evaluated without changing that basic structural representation. WSINDy-PDE
uses weak integral identities to reduce sensitivity to pointwise derivative
estimates, while DeepMoD differentiates a neural approximation of the observed
state before applying sparse regression
\citep{weak_sindy_pde_2021,deepmod_2021}. This formulation is efficient when the
supplied dictionary contains the governing terms; its structural search cannot
construct operator compositions absent from that dictionary.

\paragraph{Expandable libraries.}
A second family enlarges a bounded structural vocabulary during learning.
PDE-Net~2.0 learns differential filters and represents the nonlinear response
with a finite-depth symbolic network \citep{pde_net_2_2019}. DLGA-PDE combines
a neural state surrogate with genetic mutation and crossover, allowing terms to
be assembled even when the initial candidate library is incomplete
\citep{dlga_pde_2020}. Stepwise-DLGA adds a second stage that recovers a
spatially or temporally varying coefficient after the equation structure has
been selected \citep{stepwise_dlga_2021}. The available genes, primitive
operations, derivative order, network depth, and search budget define the
expanded hypothesis class.

\paragraph{Open-form grammars.}
Open-form methods construct complete expressions from operands and operators.
Early genetic symbolic regression demonstrated free-form searches for compact
natural laws \citep{schmidt_lipson_2009}. PDE-specific systems extend this idea
to differential operators: SGA-PDE represents an equation as a forest of
expression trees \citep{sga_pde_2022}; DISCOVER and R-DISCOVER use reinforcement
learning to
generate and refine symbolic PDE trees
\citep{discover_2024,r_discover_2024}; and EqGPT generates sequences of
operators and terms using a model trained on handbook equations \citep{eqgpt_2025}.
These representations admit structures absent from a finite term matrix, at the
cost of a combinatorial search whose reach is determined by the grammar, priors,
and search budget. We use \emph{open form} in this precise sense.

HER-PDE makes coefficient fields part of the compositional hypothesis. Thus
$c(\bm{x})\Delta u$ and
$\nabla\!\cdot\!\bigl(c(\bm{x})\nabla u\bigr)$ remain distinct hypotheses.
The complete tree retains their different operator--field structure; HEI then
estimates the declared fields and their induced spatial derivatives, which enter
linearly once the tree is fixed.

\subsection{Coefficient representations in heterogeneous systems}
\label{sec:related-work-coefficients}

Classical coefficient inverse problems begin with a known governing structure
and estimate its parameters or fields; equation discovery must select the
structure while estimating those quantities \citep{lou_review_2026}.
Constant-coefficient discovery estimates one scalar per active term, as in the
standard sparse-regression formulation of PDE-FIND
\citep{pde_find_2017}. Variable coefficients replace these scalars with
functions and couple structural selection to a function-estimation problem.
Here, coefficient regimes describe assumptions about the target field. Kernel,
spline, and neural parameterizations are estimation devices and do not by
themselves imply a compact analytic coefficient law.
Methods for compact, equation-expressible coefficients impose enough structure
to make that problem finite dimensional. Stepwise-DLGA, for example, searches
for an explicit spatial or temporal coefficient law after identifying the PDE
structure \citep{stepwise_dlga_2021}. SEQL and HEQL use symbolic neural
architectures to learn a shared equation structure with compact coefficient
expressions that vary with time, space, or another parameter
\citep{seql_heql_2023}. SGTR is a boundary method: it estimates coefficient
profiles over space or time under a shared-support assumption without requiring
a compact symbolic formula \citep{sgtr_2019}. Lou et al. place SGTR in the
equation-expressible parent category; under the finer taxonomy used here, it
spans compact and nonparametric coefficient representations. Beyond compact
coefficient laws, structural and field errors become tightly coupled. A flexible
field can absorb a missing operator, while an incorrect operator scope changes
which spatial derivatives of that field enter the equation. Joint discovery is
therefore harder than fitting scalar or low-dimensional parametric coefficients.

The regime closest to this work contains time-invariant spatial fields that are
not assumed to admit a short analytic expression. HIN-PDE combines sparse term
selection with spatial kernel regression for highly nonlinear coefficient
fields \citep{luo2023physics}; HEI builds on this spatial-kernel construction.
WG-IDENT instead uses a weak
formulation, B-spline coefficient fields, and group sparsity over a prescribed
feature dictionary \citep{wg_ident_2025}. In a domain-specific setting,
Extended-DeepGS combines group sparse regression with neural coefficient
reconstruction for water flow in continuously heterogeneous soils
\citep{song_heterogeneous_soil_2025}. These studies establish that
fixed-library terms and time-invariant spatial fields can be recovered jointly
within prescribed feature dictionaries or assumed equation forms. HER-PDE
addresses a distinct expression-level problem: each candidate determines its
operator--field structure before HEI fits the declared fields.

Nonparametric time-varying and spatiotemporal fields introduce another axis of
variation. GP-IDENT represents spatially and temporally varying coefficient
functions with B-spline bases and performs group selection from a prescribed
feature dictionary \citep{gp_ident_2023}. Bayesian group-sparse discovery has
likewise been used
for spatially or temporally varying coefficients while quantifying uncertainty
in term selection and coefficient estimates \citep{bayesian_varcoef_2021}.
These methods already occupy this regime and expose the cost of estimating a
changing field while resolving structural ambiguity.

Irregular and stochastic settings cover several different scientific tasks.
Variational-Bayes SPDE discovery estimates stochastic drift and diffusion terms
from sampled dynamics \citep{stochastic_pde_vb_2024}. Coarse-grained equation
discovery can instead learn deterministic evolution equations for probability
distributions generated by systems with random inputs or coefficients
\citep{coarse_grained_equations_2021}. Both concern stochasticity, but their
target differs from recovering one realized coefficient field shared across
experiments. Direct recovery of irregular or stochastic coefficient fields
remains comparatively sparse \citep{lou_review_2026}. The current candidate
class targets sufficiently regular, time-invariant spatial fields represented
through HEI's spatial kernels. Time-varying, spatiotemporal, and stochastic
coefficient laws lie outside its current scope.

\subsection{Language-model agents for equation discovery}
\label{sec:related-work-agents}

Language models have recently been coupled to numerical search and evaluation
for equation discovery. LLM4ED alternates equation generation with numerical
feedback and evolutionary refinement \citep{llm4ed_2024}. LLM-SR represents
equations as programs, uses an LLM to propose new skeletons, and optimizes their
parameters against data inside an evolutionary loop \citep{llm_sr_2025}.
The Scientific Generative Agent couples discrete LLM hypotheses to
differentiable simulation and observational feedback for constitutive-law and
molecular-design tasks \citep{scientific_generative_agent_2024}. For PDEs,
EqGPT combines generative structure proposals with an evaluation and
optimization loop \citep{eqgpt_2025}, and LLM-PDESR scores generated hypotheses
with subdomain weighted residuals and Pareto feedback
\citep{llm_pdesr_2026}. SR-Scientist takes a more general agentic route, writing,
executing, and revising equation programs through tools
\citep{sr_scientist_2026}.

HER-PDE applies this broader proposal, evaluation, and revision pattern to
heterogeneous PDEs. Its distinction lies in the evaluation interface: each
complete tree declares its coefficient fields, HEI fits those fields without
changing the submitted operator structure, and bidirectional transfer tests the
resulting law across two independent excitations. Field fitting and
cross-trajectory evidence therefore guide both revision and final selection.

\section{Method}
\label{sec:method}

We formulate governing-equation discovery as a closed-loop scientific inference
problem. A language-model Agent first develops a quantitative account of the
observations, then proposes complete symbolic laws and revises them using
structured numerical evidence. The Hypothesis Evaluation Interface (HEI)
connects these two activities: it fits the unknown coefficient fields declared
by a candidate and measures whether the resulting law transfers between
independent trajectories.

\subsection{Problem formulation}
\label{sec:problem-formulation}

Let $\Omega\subset\mathbb{R}^{d}$ be a spatial domain and let
$u^{(r)}:\Omega\times[0,T]\rightarrow\mathbb{R}$ denote trajectory $r$. We
observe the state on a space--time grid through
\begin{equation}
    y^{(r)}(\bm{x}_{i},t_j)
    = u^{(r)}(\bm{x}_{i},t_j) + \varepsilon^{(r)}_{ij},
    \qquad r\in\{a,b\},
    \label{eq:noisy-observations}
\end{equation}
where the two trajectories arise from different excitations of the same
spatially heterogeneous system. They therefore share a governing structure and
its time-invariant coefficient fields, while their state histories differ.

We seek a complete evolution law
\begin{equation}
    \partial_t u
    = \mathcal{F}_{h}\!\left(J^{p}u;
      c_1(\bm{x}_{S_1}),\ldots,c_{K_h}(\bm{x}_{S_{K_h}})\right),
    \label{eq:governing-law}
\end{equation}
where $J^{p}u$ is the spatial derivative jet up to order $p$. The discrete
structure $h$ specifies the expression tree: its terms, nonlinear operations,
derivative order and direction, operator scope, and the placement and reuse of
coefficient fields. It also determines the number of fields $K_h$ and each
field's spatial dependency set $S_k\subseteq\{1,\ldots,d\}$. The functions
$c_k$ are otherwise unknown and are inferred from the observations.

The admissible candidate class is nonlinear in the measured state and its
derivatives but linear in the unknown fields and their induced spatial
derivatives. This includes, for example, both
\begin{equation}
    c_0(\bm{x})\,\Delta u
    \quad\text{and}\quad
    \nabla\!\cdot\!\left(c_0(\bm{x})\nabla u\right).
    \label{eq:operator-placement-example}
\end{equation}
These are distinct hypotheses: the second expression also contains the effect
of $\nabla c_0$. We therefore represent a candidate as an expression tree and
preserve its operator placement, rather than flattening every proposal into a
fixed library of local terms. Repeated use of the same field symbol imposes one
shared spatial function; a new symbol introduces an independently fitted field.
Discovery must determine $h$ as well as the functions associated with it.

\subsection{Hypothesis Evaluation Interface}
\label{sec:hei}

HEI maps a complete symbolic hypothesis to validation results, transfer scores,
and field-level diagnostics. It first checks that the equation is a valid
evolution law in the candidate language and compiles the submitted expression
without algebraically rearranging its tree. HEI fits only the coefficient
fields explicitly declared in that expression; every term and operator in the
evaluated law originates from the Agent's hypothesis.

\paragraph{Frozen derivatives from noisy observations.}
HEI reconstructs each trajectory on its complete regular tensor grid. Let
$\bm{z}=(x_1,\ldots,x_d,t)$ denote the grid axes, with physical spacings $h_a$.
We use a separable tensor-product form of the Savitzky--Golay local-polynomial
estimator \citep{savitzky1964smoothing}. Along one axis $z_a$, an odd window of
length $w=2q+1$ is indexed by offsets $\rho=-q,\ldots,q$. A polynomial of degree
$P$ is fitted through the local samples using the Vandermonde matrix
\begin{equation}
    V_{\rho m}=\rho^m,
    \qquad \rho=-q,\ldots,q,\quad m=0,\ldots,P.
    \label{eq:savgol-vandermonde}
\end{equation}
For the window vector $\bm{y}^{(a)}_i$ centered at grid index $i$, the derivative
of order $\nu$ in physical units is
\begin{equation}
    \widehat{\partial_{z_a}^{\nu}y}_i
    = \frac{\nu!}{h_a^\nu}\,
      \bm{e}_\nu^{\mathsf T}
      (V^{\mathsf T}V)^{-1}V^{\mathsf T}\bm{y}^{(a)}_i.
    \label{eq:savgol-derivative}
\end{equation}
The case $\nu=0$ is local-polynomial smoothing. At a grid boundary, the same
degree-$P$ polynomial is fitted to the first or last complete window and
evaluated at the required boundary coordinate.

The multidimensional estimator is the tensor product of these one-dimensional
operators. Writing $W_a^{(\nu)}$ for Eq.~\eqref{eq:savgol-derivative} applied along
axis $a$, the derivative associated with multi-index $\bm{\gamma}$ is
\begin{equation}
    \widehat{D^{\bm{\gamma}}u}
    = W_{x_1}^{(\gamma_1)}\cdots
      W_{x_d}^{(\gamma_d)}W_t^{(\gamma_t)}y.
    \label{eq:tensor-savgol}
\end{equation}
For the two-dimensional experiments, HEI constructs
\begin{equation}
\begin{aligned}
    \widehat{u_t}   &=W_x^{(0)}W_y^{(0)}W_t^{(1)}y, &
    \widehat{u_x}   &=W_x^{(1)}W_y^{(0)}W_t^{(0)}y,\\
    \widehat{u_y}   &=W_x^{(0)}W_y^{(1)}W_t^{(0)}y, &
    \widehat{u_{xx}}&=W_x^{(2)}W_y^{(0)}W_t^{(0)}y,\\
    \widehat{u_{yy}}&=W_x^{(0)}W_y^{(2)}W_t^{(0)}y, &
    \widehat{u}     &=W_x^{(0)}W_y^{(0)}W_t^{(0)}y.
\end{aligned}
\label{eq:estimated-jet}
\end{equation}
All reported experiments use $w=13$ and $P=3$. Derivatives are computed before
trimming; strong-form evaluation then removes six grid points from each spatial
boundary and the first time point. Search derivatives are constructed only from
the raw temporal prefix available to search. HEI constructs the sealed-final
derivatives separately, using the final segment and the minimum preceding
context needed to fill one window. The estimator configuration and the resulting
derivative arrays are frozen before any candidate is scored:
\begin{equation}
    \widehat{J}^{p}u^{(r)}
    = \mathcal{D}_{\eta}\!\left(y^{(r)}\right).
    \label{eq:frozen-derivative-jet}
\end{equation}

\paragraph{Four-way temporal split.}
For each trajectory $r\in\{a,b\}$, the evaluated time slices are ordered and
partitioned into four contiguous intervals,
\begin{equation}
    \mathcal{I}^{(r)}
    =\mathcal{I}_{\mathrm{fit}}^{(r)}
    \mathbin{\dot\cup}\mathcal{I}_{\mathrm{dev}}^{(r)}
    \mathbin{\dot\cup}\mathcal{I}_{\mathrm{score}}^{(r)}
    \mathbin{\dot\cup}\mathcal{I}_{\mathrm{final}}^{(r)},
    \qquad
    \frac{1}{|\mathcal{I}^{(r)}|}
    \begin{pmatrix}
      |\mathcal{I}_{\mathrm{fit}}^{(r)}|\\
      |\mathcal{I}_{\mathrm{dev}}^{(r)}|\\
      |\mathcal{I}_{\mathrm{score}}^{(r)}|\\
      |\mathcal{I}_{\mathrm{final}}^{(r)}|
    \end{pmatrix}
    \simeq
    \begin{pmatrix}0.40\\0.20\\0.20\\0.20\end{pmatrix}.
    \label{eq:four-way-temporal-split}
\end{equation}
Integer time-slice counts are assigned deterministically while preserving these
proportions and keeping every interval nonempty. The first three intervals form
the search-visible prefix. The final interval remains unavailable to candidate
generation, HEI search scores, and hyperparameter selection.

\paragraph{Nonparametric coefficient fields.}
HEI builds on the spatial-kernel coefficient estimation used in HIN-PDE
\citep{luo2023physics}, adapting it to complete expression-tree hypotheses,
shared fields, analytic field derivatives, and cross-trajectory evaluation. For
field $c_k$, let $\bm{z}=\bm{x}_{S_k}$ collect its declared spatial dependencies.
HEI normalizes each coordinate to the unit interval,
\begin{equation}
    \widetilde z_j=\frac{z_j-L_{kj}}{U_{kj}-L_{kj}},
    \label{eq:coordinate-normalization}
\end{equation}
where $L_{kj}$ and $U_{kj}$ are the observed coordinate bounds. It places
anchors $\{\bm{a}_{km}\}_{m=1}^{M_k}$ on a tensor-product grid formed from
equally spaced empirical-quantile levels of the unique coordinates. The
unnormalized Gaussian response and the row-normalized kernel are
\begin{equation}
\begin{aligned}
    g_{km}(\bm{z};\ell)
    &=\exp\!\left(
      -\frac{\|\widetilde{\bm{z}}-\widetilde{\bm{a}}_{km}\|_2^2}
      {2\ell^2}\right),\\
    \kappa_{km}(\bm{z};\ell)
    &=\frac{g_{km}(\bm{z};\ell)}
      {\sum_{n=1}^{M_k}g_{kn}(\bm{z};\ell)}.
\end{aligned}
\label{eq:normalized-gaussian-kernel}
\end{equation}
The fitted field is
\begin{equation}
    c_k(\bm{z})
    = \sum_{m=1}^{M_k}\alpha_{km}\kappa_{km}(\bm{z};\ell).
    \label{eq:field-expansion}
\end{equation}
Row normalization makes the basis a partition of unity. In particular, a
spatially constant field is represented by equal coefficients at every anchor.

Operator scope can require spatial derivatives of a field. Differentiating
Eq.~\eqref{eq:normalized-gaussian-kernel} analytically gives
\begin{equation}
    \partial_{z_j}\kappa_{km}(\bm{z};\ell)
    =\frac{\kappa_{km}(\bm{z};\ell)}
      {\ell^2(U_{kj}-L_{kj})}
      \left[
      \widetilde a_{kmj}
      -\sum_{n=1}^{M_k}\kappa_{kn}(\bm{z};\ell)
       \widetilde a_{knj}
      \right].
    \label{eq:normalized-kernel-derivative}
\end{equation}
The compiler preserves the candidate tree and collects, for each field, a
pointwise multiplier $\phi_{hk}$ and derivative multipliers $\psi_{hkj}$. Its
contribution at sample $i$ has the linear form
\begin{equation}
    \phi_{hk,i}c_k(\bm{z}_i)
    +\sum_{j\in S_k}\psi_{hkj,i}\partial_{z_j}c_k(\bm{z}_i).
    \label{eq:field-effect}
\end{equation}
For example, $\partial_x(c_k u_x)$ yields
$\phi_{hk}=u_{xx}$ and $\psi_{hkx}=u_x$. The resulting design block is
\begin{equation}
    B_{hk,im}
    =\phi_{hk,i}\kappa_{km}(\bm{z}_i;\ell)
    +\sum_{j\in S_k}\psi_{hkj,i}
      \partial_{z_j}\kappa_{km}(\bm{z}_i;\ell).
    \label{eq:field-design-block}
\end{equation}
All fields in a candidate are concatenated into one design matrix
$B_h=[B_{h1}\mid\cdots\mid B_{hK_h}]$ and fitted jointly. Let
$\bm{r}_h=\widehat{\bm{u}}_t-\bm{f}_{h,0}$ be the time derivative after
subtracting any field-free part of the candidate. HEI scales each design column
by its root-mean-square magnitude,
\begin{equation}
    s_m=\left(\frac{1}{n}\sum_{i=1}^{n}B_{h,im}^2\right)^{1/2},
    \qquad
    \overline B_h=B_h\operatorname{diag}(\bm{s})^{-1},
    \label{eq:design-normalization}
\end{equation}
using $s_m=1$ for a zero column. It then solves
\begin{equation}
    \widehat{\bm{\beta}}^{(r)}_{h,\ell,\lambda}
    =\arg\min_{\bm{\beta}}
      \frac{1}{n}\|\bm{r}_h-\overline B_h\bm{\beta}\|_2^2
      +\lambda\|\bm{\beta}\|_2^2,
    \qquad
    \widehat{\bm{\alpha}}
    =\operatorname{diag}(\bm{s})^{-1}\widehat{\bm{\beta}}.
    \label{eq:field-fit}
\end{equation}
For each transfer direction $r\rightarrow s$, HEI fits on
$\mathcal{I}_{\mathrm{fit}}^{(r)}$ and selects the kernel width $\ell$ and
regularization $\lambda$ on $\mathcal{I}_{\mathrm{dev}}^{(r)}$. It then refits
the fields on the union of those two source intervals and evaluates the fitted
law on $\mathcal{I}_{\mathrm{score}}^{(s)}$ from the other trajectory.

For a transfer from source interval $\mathcal{I}_{\mathrm{tr}}^{(r)}$ to target
interval $\mathcal{I}_{\mathrm{te}}^{(s)}$, the normalized prediction error is
\begin{equation}
    E_{r\rightarrow s}
      \!\left(h;\mathcal{I}_{\mathrm{tr}}^{(r)},
      \mathcal{I}_{\mathrm{te}}^{(s)}\right)
    =
    \frac{
      \left\|\widehat{u_t}^{(s)}_{\mathcal{I}_{\mathrm{te}}}
      - \widehat{\mathcal{F}}_{h}^{(r\rightarrow s)}
        \big|_{\mathcal{I}_{\mathrm{te}}}\right\|_2^2
    }{
      \left\|\widehat{u_t}^{(s)}_{\mathcal{I}_{\mathrm{te}}}
      - \overline{\widehat{u_t}^{(s)}_{\mathcal{I}_{\mathrm{te}}}}\right\|_2^2
      + \delta\max\!\left(
        1,\left\|\widehat{u_t}^{(s)}_{\mathcal{I}_{\mathrm{te}}}\right\|_2^2
      \right)
    },
    \label{eq:transfer-error}
\end{equation}
where the coefficient fields in
$\widehat{\mathcal{F}}_{h}^{(r\rightarrow s)}$ are fitted only on
$\mathcal{I}_{\mathrm{tr}}^{(r)}$, and $\delta$ is a numerical floor. The
search-visible HEI score is
\begin{equation}
\begin{aligned}
    E_{\mathrm{search}}(h)=\frac{1}{2}\bigl[&
      E_{a\rightarrow b}\!\left(
        h;\mathcal{I}_{\mathrm{fit}}^{(a)}\!\cup
        \mathcal{I}_{\mathrm{dev}}^{(a)},
        \mathcal{I}_{\mathrm{score}}^{(b)}
      \right)\\
      +{}&E_{b\rightarrow a}\!\left(
        h;\mathcal{I}_{\mathrm{fit}}^{(b)}\!\cup
        \mathcal{I}_{\mathrm{dev}}^{(b)},
        \mathcal{I}_{\mathrm{score}}^{(a)}
      \right)\bigr].
\end{aligned}
\label{eq:hei-score}
\end{equation}
Lower values indicate better transfer. Symmetric transfer rewards a coefficient
field that explains both experiments instead of a flexible function that
interpolates one trajectory. HEI ranks valid candidates by exact
$E_{\mathrm{search}}$; exact ties are ordered by right-hand-side term count,
expression-tree size, and canonical representation.

After the Agent selects $\widehat{h}$, HEI keeps the search-selected
hyperparameters fixed, refits each direction on its first three intervals, and
evaluates the other trajectory's final interval:
\begin{equation}
\begin{aligned}
    E_{\mathrm{sealed}}(\widehat{h})=\frac{1}{2}\bigl[&
      E_{a\rightarrow b}\!\left(
        \widehat{h};\mathcal{I}^{(a)}\!\setminus
        \mathcal{I}_{\mathrm{final}}^{(a)},
        \mathcal{I}_{\mathrm{final}}^{(b)}
      \right)\\
      +{}&E_{b\rightarrow a}\!\left(
        \widehat{h};\mathcal{I}^{(b)}\!\setminus
        \mathcal{I}_{\mathrm{final}}^{(b)},
        \mathcal{I}_{\mathrm{final}}^{(a)}
      \right)\bigr].
\end{aligned}
\label{eq:hei-sealed-score}
\end{equation}
The final $20\%$ therefore provides a temporal audit that never influences the
search trajectory or final structural choice.

The interface returns the canonical equation, validity status, forward and
reverse transfer errors, archive rank, and evaluation-budget counters. For a
valid field-bearing candidate, it can also expose the two independently fitted
fields on common spatial anchors together with design conditioning and
cross-trajectory stability. These diagnostics let the Agent test whether two
apparent terms support a shared field, whether an added field is identifiable,
and whether a compact operator form explains an expanded local relation. The
ground-truth equation, true coefficient fields, and sealed final score remain
outside the search interface.

\paragraph{Agent-visible HEI interaction protocol.}
The Agent accesses HEI through a finite action set exposed by the \texttt{pde}
terminal front end,
\begin{equation}
    \mathcal{U}_{\mathrm{HEI}}
    =\{\mathsf{status},\mathsf{validate},\mathsf{score},
      \mathsf{score\mbox{-}batch},\mathsf{history},
      \mathsf{evidence},\mathsf{submit}\}.
    \label{eq:hei-action-set}
\end{equation}
Every action returns one versioned structured record. Table~\ref{tab:hei-interface}
defines the scientific semantics of these actions; Appendix~\ref{app:dsl-prompt}
reproduces their exact terminal syntax together with the candidate language.

\begin{table}[!ht]
  \centering
  \caption{Agent-visible HEI interaction protocol. A unique evaluation is
  charged only when HEI scores a previously unseen canonical candidate.}
  \label{tab:hei-interface}
  \vspace{3pt}
  \setlength{\tabcolsep}{3pt}
  \renewcommand{\arraystretch}{1.15}
  \footnotesize
  \begin{tabular}{@{}>{\raggedright\arraybackslash}p{0.22\linewidth}
                    >{\raggedright\arraybackslash}p{0.65\linewidth}
                    >{\raggedright\arraybackslash}p{0.09\linewidth}@{}}
    \toprule
    Command & Agent-visible operation and response & Cost \\
    \midrule
    \texttt{pde status}
      & Reports the unique-evaluation budget, submission state, and current
        archive winner. & None \\
    \texttt{pde validate <file>}
      & Parses a complete candidate and checks the candidate-language and
        evolution-law constraints without fitting fields. & None \\
    \texttt{pde score <file>}
      & Canonicalizes and evaluates one complete candidate. The response
        contains its identifier, validity or stable failure code, directional
        transfer errors, symmetric score, rank, cache state, and counters. &
        \texttt{1/new} \\
    \texttt{pde score-batch <files>}
      & Applies \texttt{score} independently to up to 32 candidates; batching
        changes throughput but not candidate semantics. & \texttt{1/new} \\
    \texttt{pde history}
      & Returns the scored archive in chronological or ranked order without
        creating a new evaluation. & None \\
    \texttt{pde evidence <id>}
      & Exports the candidate's already-fitted coefficient fields on common
        anchors, together with conditioning, selected hyperparameters, and
        cross-trajectory stability diagnostics. & None \\
    \texttt{pde submit <id>}
      & Selects an already-scored valid candidate and terminates Agent search.
        HEI performs sealed-final evaluation only after this selection. & None \\
    \bottomrule
  \end{tabular}
\end{table}

Canonical duplicates return their cached records and leave the unique-score
budget unchanged. Field evidence reports numerical properties of a scored
structure; it supplies no structural verdict. None of the actions exposes the
generating equation, true fields, operator-equivalence labels, or sealed-final
score during search.

\subsection{Two-phase scientific-agent workflow}
\label{sec:agent-workflow}

\begin{figure}[!ht]
    \centering
    \includegraphics[width=0.74\linewidth]{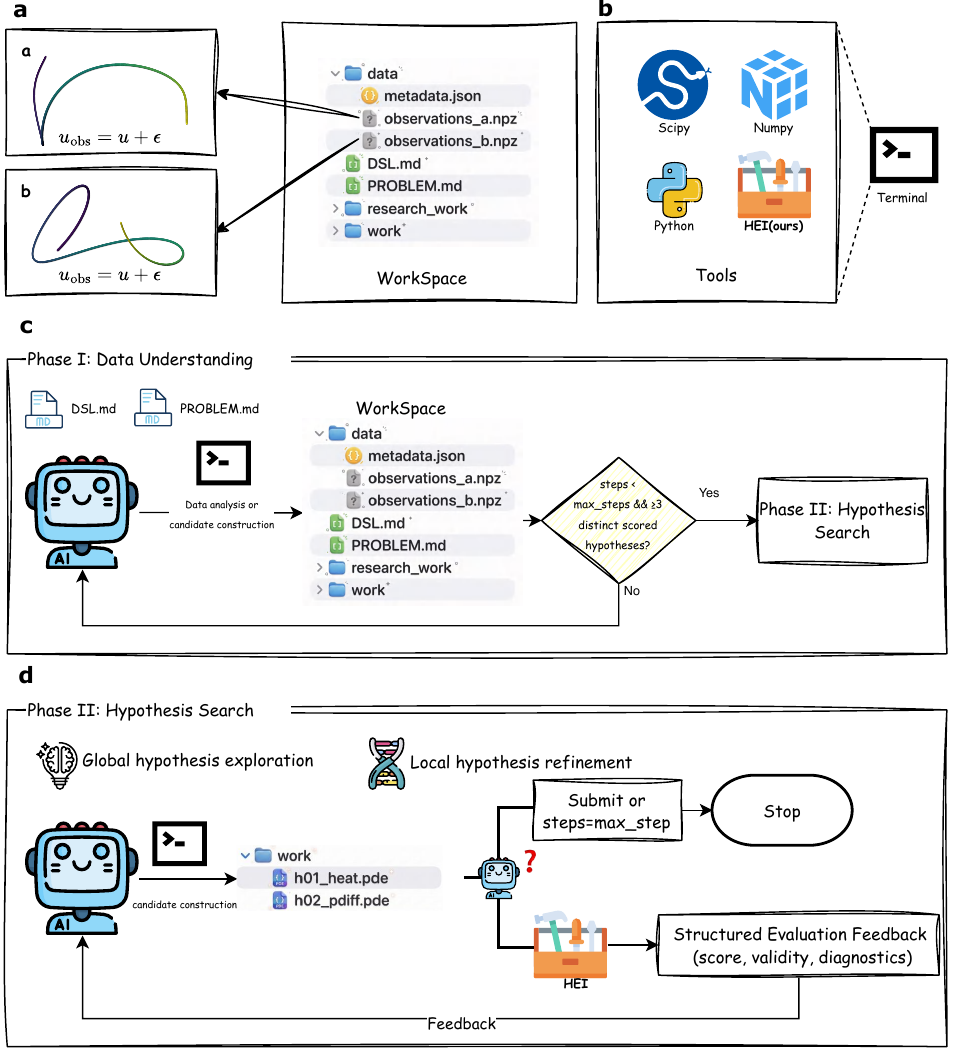}
    \caption{Overview of the scientific-agent workflow. \textbf{(a)} The
    workspace contains two noisy trajectories, experimental context, the
    candidate language, and persistent research files. \textbf{(b)} The Agent
    drives numerical tools and HEI through a common terminal interface.
    \textbf{(c)} During Phase~I, it analyzes the observations and constructs an
    initial archive of at least three distinct HEI-scored hypotheses.
    \textbf{(d)} During Phase~II, global hypothesis exploration and local
    hypothesis refinement generate complete candidates; HEI returns structured
    evaluation feedback until the Agent submits a final hypothesis or exhausts
    the search budget.}
    \label{fig:agent-hei-workflow}
\end{figure}

Figure~\ref{fig:agent-hei-workflow} shows how the Agent combines open-ended data
analysis with controlled hypothesis search. The workflow maintains one
conversation across both phases, so numerical findings, rejected explanations,
and scientific reasoning remain available when the search policy changes.

\paragraph{Controller state and scored archive.}
Let $\tau$ index Agent turns, distinct from the physical time $t$ in
Eq.~\eqref{eq:noisy-observations}. After turn $\tau$, the controller state is
\begin{equation}
    \mathsf{S}_{\tau}
    =\left(
      \mathcal{C}_{\tau},
      \mathcal{W}_{\tau},
      \mathcal{A}_{\tau},
      \varphi_{\tau},
      \mathcal{R}_{\tau}
    \right),
    \label{eq:agent-controller-state}
\end{equation}
where $\mathcal{C}_{\tau}$ is the complete conversation, $\mathcal{W}_{\tau}$
is the visible workspace, $\mathcal{A}_{\tau}$ is the scored candidate archive,
$\varphi_{\tau}$ is the controller mode, and $\mathcal{R}_{\tau}$ records the
remaining turn, evaluation, and wall-time budgets.

Write $\xi$ for a complete candidate equation and $\operatorname{can}(\xi)$
for its canonical representation. Evaluating a candidate produces the
structured record
\begin{equation}
    \mathsf{Z}(\xi)=\operatorname{HEI}(\xi),
    \label{eq:agent-hei-feedback}
\end{equation}
which contains its validity status, canonical equation, two transfer errors,
symmetric score, archive rank, and budget counters; fitted-field evidence can
be requested for a valid field-bearing candidate. The archive is keyed by the
canonical representation,
\begin{equation}
    \mathcal{A}_{\tau}
    =\left\{
      \bigl(\operatorname{can}(\xi),\mathsf{Z}(\xi)\bigr)
      : \xi\ \text{has received a unique HEI evaluation by turn }\tau
    \right\}.
    \label{eq:scored-candidate-archive}
\end{equation}
A cached duplicate therefore leaves $\mathcal{A}_{\tau}$ unchanged. Candidates
that pass the language parser but fail numerical evaluation remain in the
archive with their failure status; only valid entries are ranked or eligible
for final submission.

\paragraph{Phase I: data understanding and checkpoint.}
The Agent begins with the experimental description, measurement metadata,
candidate-language specification, and both raw trajectories. It uses Python,
NumPy, and SciPy to reconstruct the measurement grid, characterize noise and
sampling, estimate informative quantities, compare broad local relations, and
test whether inferred quantities remain consistent across trajectories and
reasonable analysis choices. It may call HEI whenever a complete candidate is
ready.

Let $\tau_{\mathrm{obs}}=16$ be the observation-analysis checkpoint and let
$N_{\min}=3$ be the minimum number of distinct scored candidates. The
phase-transition index is determined by
\begin{equation}
    \tau_{\star}
    =\min\left\{
      \tau:\tau\geq\tau_{\mathrm{obs}}
      \ \land\ |\mathcal{A}_{\tau}|\geq N_{\min}
    \right\},
    \qquad
    \varphi_{\tau}=
    \begin{cases}
        \mathrm{I},  & \tau<\tau_{\star},\\
        \mathrm{II}, & \tau\geq\tau_{\star}.
    \end{cases}
    \label{eq:phase-transition}
\end{equation}
with $\tau_{\star}=\infty$ when the gate is not reached within the exploration
budget. Because $\mathsf{S}_{\tau}$ denotes the state after turn $\tau$, a gate
reached at $\tau=16$ governs the next Agent action.
If fewer than $N_{\min}$ candidates have been scored after turn
$\tau_{\mathrm{obs}}$, the controller admits only candidate construction and
scoring actions until the archive reaches the threshold. The checkpoint
converts the initial data analysis into several explicit, competing
explanations.

The transition preserves the reasoning history and changes the available
workspace:
\begin{equation}
    \mathcal{C}^{\mathrm{II}}_{\tau_{\star}}
      =\mathcal{C}^{\mathrm{I}}_{\tau_{\star}},
    \qquad
    \mathcal{A}^{\mathrm{II}}_{\tau_{\star}}
      =\mathcal{A}^{\mathrm{I}}_{\tau_{\star}},
    \qquad
    \mathcal{W}^{\mathrm{II}}_{\tau_{\star}}
      =\operatorname{Freeze}\!\left(
        \mathcal{W}^{\mathrm{I}}_{\tau_{\star}}
      \right).
    \label{eq:workspace-freeze}
\end{equation}
$\operatorname{Freeze}$ removes the raw observation arrays and research-phase
caches. It retains the problem and language specifications, candidate files,
HEI, the scored archive, exported fitted-field evidence, and terminal-driven
numerical tools. Phase~II consequently begins from the same scientific account
of the data without reopening raw-data analysis.

\paragraph{Phase II: dual-channel hypothesis search.}
Phase~II supplies two complementary candidate-construction operators. Global
hypothesis exploration produces
\begin{equation}
    \xi^{\mathrm{glob}}_{\tau+1}
    =\mathfrak{G}\!\left(
      \mathcal{C}_{\tau},\mathcal{W}_{\tau},\mathcal{A}_{\tau}
    \right),
    \label{eq:global-construction}
\end{equation}
by returning to the accumulated evidence and building a complete expression tree
without using a scored candidate as its template. It searches materially
different operator compositions, field identities and dependencies, nonlinear
couplings, derivative placements, and other structural axes exposed by the
candidate language. Local hypothesis refinement produces
\begin{equation}
    \xi^{\mathrm{loc}}_{\tau+1}
    =\mathfrak{L}\!\left(
      \xi^{\mathrm{par}},\mathsf{Z}(\xi^{\mathrm{par}});
      \mathcal{C}_{\tau},\mathcal{W}_{\tau},\mathcal{A}_{\tau}
    \right),
    \label{eq:local-revision}
\end{equation}
by selecting a scored parent $\xi^{\mathrm{par}}$, stating one evidence-based
weakness or ambiguity, and applying one controlled structural change. The Phase-II policy
instructs the Agent to contribute scored proposals from both operators while
leaving their timing and scientific content to the Agent.

For either construction channel, one search iteration has the update
\begin{equation}
\begin{aligned}
    \xi_{\tau+1}
      &\in\left\{
        \xi^{\mathrm{glob}}_{\tau+1},
        \xi^{\mathrm{loc}}_{\tau+1}
      \right\},\\
    \mathsf{Z}_{\tau+1}
      &=\operatorname{HEI}(\xi_{\tau+1}),\\
    \mathcal{A}_{\tau+1}
      &=
      \begin{cases}
        \mathcal{A}_{\tau}\cup
        \left\{\bigl(
          \operatorname{can}(\xi_{\tau+1}),\mathsf{Z}_{\tau+1}
        \bigr)\right\},
        & \operatorname{can}(\xi_{\tau+1})
          \notin\operatorname{dom}(\mathcal{A}_{\tau}),\\
        \mathcal{A}_{\tau}, & \text{otherwise}.
      \end{cases}
\end{aligned}
\label{eq:hypothesis-search-update}
\end{equation}
Here membership in $\mathcal{A}_{\tau}$ is tested by canonical equation rather
than by source filename. The HEI record is appended to the same conversation,
so numerical feedback and field evidence can determine the next construction.

\paragraph{Termination and scientific selection.}
An explicit submission terminates exploration immediately. If the exploration
budget ends first, the controller enters a bounded selection mode in which the
only permitted operations are archive inspection and submission of an already
scored valid candidate. With $\tau_{\mathrm{end}}$ denoting the final
exploration turn and $\mathcal{A}^{+}_{\tau_{\mathrm{end}}}$ its valid archive,
the reported structure is
\begin{equation}
    \widehat{\xi}
    =\operatorname{Select}_{\mathrm{Agent}}\!\left(
      \mathcal{C}_{\tau_{\mathrm{end}}},
      \mathcal{A}^{+}_{\tau_{\mathrm{end}}}
    \right),
    \qquad
    \widehat{\xi}
      \in\operatorname{dom}(\mathcal{A}^{+}_{\tau_{\mathrm{end}}}).
    \label{eq:agent-selection}
\end{equation}
The Agent weighs HEI rank together with structural coherence, field evidence,
and unsupported complexity. HEI then refits the fields declared by
$\widehat{\xi}$ using all search-visible time windows, keeps the selected kernel
and regularization hyperparameters fixed, and reports symmetric transfer on the
sealed final windows.

\section{Experiments}
\label{sec:experiments}

\subsection{Experimental protocol}
\label{sec:experimental-protocol}

We study five two-dimensional systems that separate complementary structural
questions: a nonconservative variable diffusivity, a shared-field conservative
flux, conservative diffusion with an independent reaction field, diffusion
with two independent transport fields, and orthotropic conservative diffusion
with two directional fields. The groundwater case uses the heterogeneous
conductivity realization from the HIN-PDE benchmark \citep{luo2023physics}.
The other four systems use non-analytic Mat\'ern-$3/2$ coefficient-field
realizations rather than fields drawn from a short analytic expression family.

Every trajectory is supplied to the Agent as raw coordinates, times, and state
measurements with $5\%$ relative Gaussian noise. Clean states, simulator
derivatives, coefficient fields, generating equations, PDE-family names, and
oracle candidates remain hidden. Every discovery run uses exactly two noisy
search trajectories, denoted A and B throughout. This notation is independent
of internal dataset labels. In every case, the derivatives seen by HEI are
estimated from the noisy state arrays and divided according to the frozen
$40\%/20\%/20\%/20\%$ temporal protocol in Section~\ref{sec:hei}.

All Agent runs use \texttt{deepseek-v4-flash} with thinking enabled,
temperature $0.7$, and high reasoning effort. Phase~I ends at turn 16 after at
least three distinct candidates have been scored. The complete rollout permits
36 model turns, two bounded selection turns, and at most 96 unique HEI
evaluations. The four synthetic systems fit each two-dimensional field on a
$10\times10$ anchor grid and select a normalized Gaussian-kernel width from
$\{0.08,0.16,0.24\}$. Groundwater uses a $12\times12$ grid and widths
$\{0.04,0.08,0.16\}$. Every case uses ridge parameter $10^{-8}$. These
settings are fixed before the Agent trajectory begins.

We assess three separate outcomes. \emph{Structural recovery} requires the
submitted expression to equal the generating operator after field renaming,
an unrestricted field-sign reparameterization, or an exact product-rule
expansion. \emph{Transfer error} comprises the search and sealed HEI errors
defined by Eqs.~\eqref{eq:hei-score} and~\eqref{eq:hei-sealed-score}.
\emph{Field recovery} compares each fitted
field with its generating field at the same HEI anchors using Pearson
correlation and relative error
\begin{equation}
    E_{c}^{\mathrm{rel}}
    =\frac{\|\widehat{c}-c\|_2}{\|c\|_2}.
    \label{eq:coefficient-relative-error}
\end{equation}

\begin{table}[t]
  \centering
  \caption{Five registered Agent discoveries. Search reports the HEI score on
  the third temporal interval; sealed reports the post-submission score on the
  final interval. The Agent-selected equations are the submitted
  structures rendered in mathematical notation.}
  \label{tab:five-discoveries}
  \vspace{3pt}
  \setlength{\tabcolsep}{3pt}
  \renewcommand{\arraystretch}{1.22}
  \footnotesize
  \begin{tabular}{@{}p{0.12\linewidth}p{0.19\linewidth}p{0.24\linewidth}p{0.19\linewidth}rr@{}}
    \toprule
    Case & Generating law & Agent-selected law & Structural relation & Search & Sealed \\
    \midrule
    Groundwater flow
      & $h_t=\nabla\!\cdot\!\left((K/S_s)\nabla h\right)$
      & $\begin{aligned}h_t={}&\partial_x(C_0h_x)\\&+\partial_y(C_0h_y)\end{aligned}$
      & Exact; $C_0=K/S_s$
      & 0.5140 & 0.1226 \\
    Variable diffusion
      & $u_t=a\Delta u$
      & $u_t=C_0(u_{xx}+u_{yy})$
      & Exact; $C_0=a$
      & 0.2093 & 0.2890 \\
    Reaction--diffusion
      & $u_t=\nabla\!\cdot(D\nabla u)-ku$
      & $\begin{aligned}u_t={}&C_{0,x}u_x+C_0u_{xx}\\&+C_{0,y}u_y+C_0u_{yy}\\&+C_1u\end{aligned}$
      & Product-rule equivalent; $C_0=D$, $C_1=-k$
      & 0.4506 & 0.4272 \\
    Advection--diffusion
      & $u_t=D\Delta u-\bm{v}\!\cdot\nabla u$
      & $\begin{aligned}u_t={}&C_0(u_{xx}+u_{yy})\\&+C_1u_x+C_2u_y\end{aligned}$
      & Signed-field equivalent; $C_0=D$, $(C_1,C_2)=-\bm{v}$
      & 0.4944 & 0.5114 \\
    Orthotropic diffusion
      & $\begin{aligned}u_t={}&\partial_x(D_xu_x)\\&+\partial_y(D_yu_y)\end{aligned}$
      & $\begin{aligned}u_t={}&\partial_x(C_0u_x)\\&+\partial_y(C_1u_y)\end{aligned}$
      & Exact; $(C_0,C_1)=(D_x,D_y)$
      & 0.2041 & 0.2761 \\
    \bottomrule
  \end{tabular}
\end{table}

\subsection{Governing-structure discovery}
\label{sec:governing-structure-results}

The Agent submits the generating structure in all five registered cases
(Table~\ref{tab:five-discoveries}). Variable diffusion preserves one shared
field across $u_{xx}$ and $u_{yy}$. Reaction--diffusion is submitted as the
five-term product-rule expansion of
$\nabla\!\cdot(C_0\nabla u)+C_1u$, retaining a single diffusivity field across
both axes. Advection--diffusion recovers one shared diffusivity and two
independent signed transport fields. Groundwater flow and orthotropic diffusion
place the unknown fields inside the corresponding spatial derivatives, thereby
recovering the conservative operator rather than only its second-derivative
component.

\begin{figure}[!ht]
    \centering
    \includegraphics[width=0.94\linewidth]{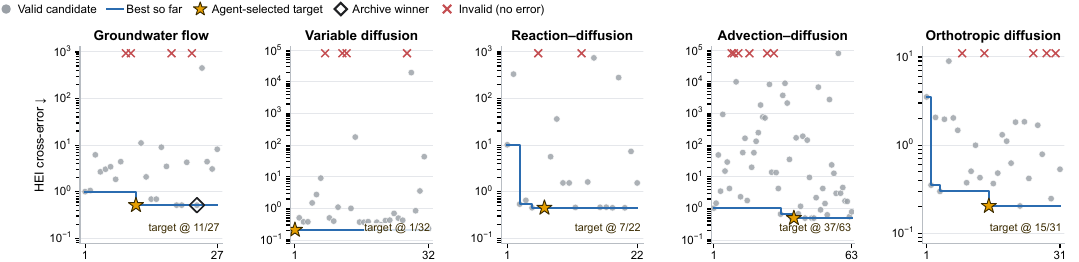}
    \caption{Agent search trajectories for the five positive discovery cases.
    Gray points are valid candidates, the blue line is the best-so-far HEI
    error, red crosses are invalid evaluations without an imputed score, and
    the gold star is the first generating-structure candidate ultimately
    submitted by the Agent. For Groundwater, the hollow diamond is the
    lower-search-error archive winner with an added unit $y$-gradient term
    (its $x$-gradient multiplier is zero); the Agent instead selects the
    generating conservative operator. Target labels are assigned
    retrospectively and are not visible during search.}
    \label{fig:five-discovery-search}
\end{figure}

Figure~\ref{fig:five-discovery-search} shows the complete HEI evaluation
sequence. The first generating-structure candidate appears at evaluations
$11/27$, $1/32$, $7/22$, $37/63$, and $15/31$ for groundwater flow, variable
diffusion, reaction--diffusion, advection--diffusion, and orthotropic diffusion,
respectively. The Agent retains each of these candidates as its final
submission while continuing to test alternative structures when budget remains.
The trajectory is not a monotone numerical optimization: later proposals probe
different operator placements, dependencies, and extra terms, and several are
invalid under the candidate language or numerical checks.
Appendix~\ref{app:agent-decision-traces} reconstructs the corresponding
decision traces from the chronological conversations and HEI candidate ledgers.
Appendix~\ref{app:top-five-candidates} reports the five lowest-error
structurally distinct valid candidates in each complete archive and marks every
expression that is operator-equivalent to the generating law.

Groundwater illustrates the difference between HEI ranking and scientific
selection. The lowest search error, $0.51318$, belongs to
\begin{equation}
    h_t=\nabla\!\cdot(C_0\nabla h)
        +0h_x+1h_y,
    \label{eq:groundwater-overfit-candidate}
\end{equation}
as written by the Agent, with the candidate-language integer literals
\texttt{N0} and \texttt{N1} rendered as $0$ and $1$. The candidate therefore
adds one effective $h_y$ term. The generating conservative
operator scores $0.51398$. The
Agent rejects the unsupported gradient term and submits the latter. On sealed
windows, the compact submission also has the slightly smaller error
($0.122599$ versus $0.122641$). Thus the final law is selected from the scored
archive using numerical transfer together with operator coherence and
unsupported complexity, rather than by returning the smallest search-visible
number mechanically.

\subsection{Unknown-field recovery}
\label{sec:unknown-field-results}

\begin{figure}[!ht]
    \centering
    \includegraphics[width=0.80\linewidth]{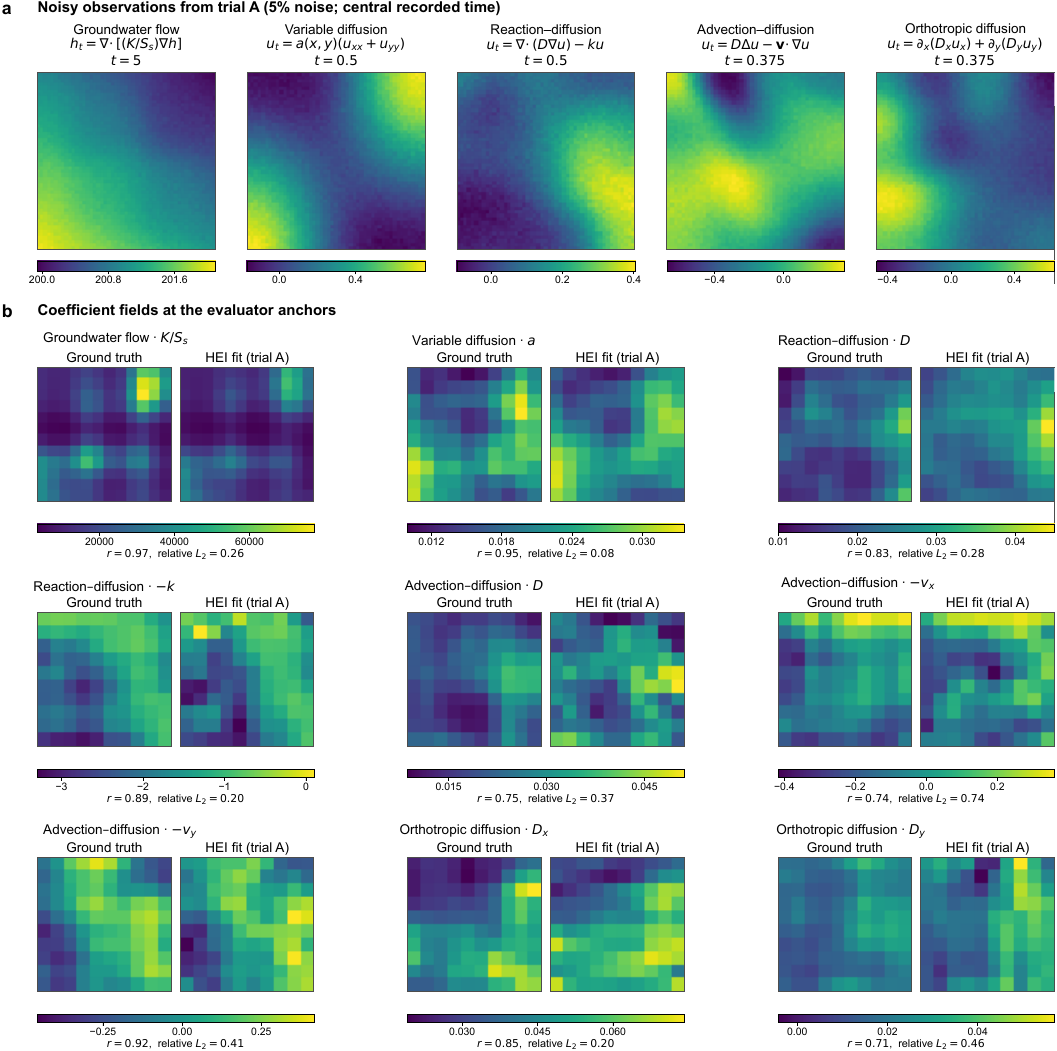}
    \caption{Noisy observations and recovered coefficient fields. \textbf{(a)}
    The central recorded time slice from trajectory A for each case; all state
    observations contain $5\%$ relative noise. \textbf{(b)} Ground truth and
    the HEI fit induced by the Agent-submitted law for all nine coefficient
    fields. Truth and fit are evaluated at identical anchors and displayed
    without spatial smoothing; every pair shares its color scale. Here $r$ is
    Pearson correlation and $E_{\mathrm{rel}}$ is defined by
    Eq.~\eqref{eq:coefficient-relative-error}.}
    \label{fig:five-discovery-fields}
\end{figure}

The recovered structures contain nine unknown two-dimensional fields in total.
Figure~\ref{fig:five-discovery-fields} compares every fitted field with its
generating field at the same HEI anchors. Pearson correlation ranges from
$0.711$ to $0.966$ across the nine fields, while relative $L_2$ error ranges
from $0.082$ to $0.736$. The variable diffusivity is recovered with
$r=0.946$ and $E_c^{\mathrm{rel}}=0.082$. The Groundwater field $K/S_s$ reaches
$r=0.966$, and both reaction--diffusion fields exceed $r=0.82$. The largest
amplitude error occurs for the advection field $-v_x$
($E_c^{\mathrm{rel}}=0.736$), whose spatial pattern remains positively
correlated with the truth ($r=0.740$).

For the reaction and advection systems, Figure~\ref{fig:five-discovery-fields}
uses the signed parameterization of the submitted equation: $C_1=-k$ for
reaction--diffusion and $(C_1,C_2)=(-v_x,-v_y)$ for
advection--diffusion. The Groundwater comparison uses $C_0=K/S_s$, matching the
normalized evolution equation rather than unscaled conductivity. Each
truth/fit pair shares one color scale, so the panels show both spatial
morphology and amplitude mismatch.

\subsection{Evidence provenance}
\label{sec:evidence-provenance}

These experiments establish five positive discovery case studies; they do not
estimate a population discovery rate. Variable diffusion and
advection--diffusion use frozen admitted benchmarks followed by one authorized
Agent trajectory. Groundwater is a controlled method-development reproduction
on the same data as an earlier successful trajectory. Reaction--diffusion uses
a trajectory pair selected during data-protocol development after an earlier
pair admitted a nonlinear shortcut.
Orthotropic diffusion uses new field, preparation, and noise seeds after a
benchmark-neutral clarification of recursive field operations in the candidate
language. Population-level recovery rates require the finalized protocol to be
run on additional unseen field realizations, preparations, noise seeds, and
Agent samples.

\section{Conclusion}
\label{sec:conclusion}

We presented an agentic framework for discovering PDE structures whose
coefficients are unknown spatial fields. A scientific Agent constructs complete
symbolic hypotheses in a language that preserves operator scope, spatial
dependencies, and shared-field identity. HEI converts each hypothesis into
cross-trajectory evidence by estimating its declared coefficient fields
nonparametrically and measuring transfer between independent experiments. The
two-stage workflow combines open-ended analysis with both creative hypothesis
generation and local candidate refinement.

Across five two-dimensional systems observed with $5\%$ state noise, the Agent
recovered the generating operator structure for groundwater flow, variable
diffusion, reaction--diffusion, advection--diffusion, and orthotropic diffusion.
These cases require distinctions between conservative and nonconservative
placement, shared and independent fields, and diffusion, transport, and
reaction mechanisms. The decision traces also show that discovery is more than
numerical ranking: the Agent uses HEI scores together with structural coherence
to select a governing law from competing, fully evaluated hypotheses. This
combination extends equation discovery from finite-dimensional symbolic forms
to structured laws with nonparametric spatial heterogeneity.

\bibliographystyle{plainnat}
\bibliography{main}

\appendix

\section{Agent Decision Traces}
\label{app:agent-decision-traces}

This appendix condenses the chronological Agent conversations and HEI candidate
ledgers into auditable decision traces. Scored equations and errors are copied
from the candidate ledgers; validation outcomes come from the chronological
conversation; and the update column paraphrases the Agent's contemporaneous
rationale. We write
$\Delta u=u_{xx}+u_{yy}$ and suppress the arguments of unrestricted
two-dimensional fields, so $C_j=C_j(x,y)$ unless stated otherwise. Lower
$E_{\mathrm{search}}$ is better.

\subsection{Detailed trajectory: advection--diffusion}
\label{app:advection-decision-trace}

During Phase~I, the Agent reconstructed a $49\times49$ spatial grid with 33
visible times per trajectory and compared spatial and temporal variation across
the two noisy records. Its exploratory finite differences showed weak
correlation between $u_t$ and $\Delta u$. A stronger empirical correlation
between $u_{tt}$ and $\Delta u$ temporarily led it toward a wave-like account.
HEI rejected that proposal because the candidate language admits one measured
first-order time rate on the left and no time derivative on the right. The
Agent then returned to first-order diffusion, transport, and reaction
hypotheses. Thus Phase~I produced competing structural directions rather than a
single fitted law.

At the Phase~II transition, the controller removed the raw arrays while
preserving the conversation and scored archive. The resulting search combined
global constructions with controlled removal, dependency, and operator-scope
tests. Table~\ref{tab:advection-decision-trace} records the decisive sequence.

\setlength\LTleft{0pt}
\setlength\LTright{0pt}
\begin{longtable}{@{}p{0.11\linewidth}p{0.33\linewidth}p{0.11\linewidth}p{0.37\linewidth}@{}}
  \caption{Detailed advection--diffusion decision trace. ``Eval.'' is the
  chronological index of a distinct HEI-scored candidate.}
  \label{tab:advection-decision-trace}\\
  \toprule
  Stage & Candidate or test & $E_{\mathrm{search}}$ & Agent update \\
  \midrule
  \endfirsthead
  \multicolumn{4}{l}{\footnotesize\emph{Table~\ref{tab:advection-decision-trace} continued}}\\
  \toprule
  Stage & Candidate or test & $E_{\mathrm{search}}$ & Agent update \\
  \midrule
  \endhead
  \bottomrule
  \endfoot
  Phase~I
    & Exploratory derivatives and cross-trial statistics
    & ---
    & Enumerated diffusion, wave, advection, and reaction--diffusion as
      competing explanations. \\
  Phase~I
    & $u_{tt}=0\Delta u$
    & invalid
    & Used HEI's language failure to rule out the second-order-time branch and
      restrict subsequent proposals to measured $u_t$. \\
  Eval.~1
    & $u_t=0\Delta u$
    & 1.0246
    & The Agent-authored zero multiplier makes this the effective null law
      $u_t=0$, establishing the first numerical baseline. \\
  Eval.~2
    & $u_t=C_0\Delta u$
    & 1.4641
    & Found that spatially varying diffusivity alone did not explain transfer
      between the two preparations. \\
  Eval.~8
    & $u_t=C_1u_x+C_2u_y$
    & 4.2552
    & Ruled out heterogeneous transport without a second-order component. \\
  Eval.~31
    & $u_t=C_0\Delta u+C_1u_x+C_2u_y+C_3u$
    & 0.6588
    & The first joint diffusion--transport construction produced a large score
      improvement; the remaining question was whether reaction was necessary. \\
  Eval.~37
    & $u_t=C_0\Delta u+C_1u_x+C_2u_y$
    & 0.4944
    & Removing reaction improved transfer and created the eventual archive
      winner. \\
  Eval.~40
    & $u_t=\nabla\!\cdot(C_0\nabla u)$
    & 1.1669
    & A global operator-placement test showed that conservative diffusion alone
      could not replace the independent transport fields. \\
  Eval.~49
    & $u_t=C_0\Delta u+\nabla C_1\!\cdot\nabla u$
    & 0.9015
    & A single potential-like field could not account for both transport
      components. \\
  Eval.~52
    & $u_t=C_0\Delta u+C_1u_x+C_2u_y+C_3\sin u$
    & 0.6258
    & A nonlinear reaction extension degraded cross-trajectory transfer. \\
  60/61
    & $C_0\Delta u+C_1u_x$; $C_0\Delta u+C_1u_y$
    & 1.7138; 1.6633
    & Directional ablations showed that both independent transport components
      were required. \\
  Eval.~62
    & $u_t=\nabla\!\cdot(C_0\nabla u)+C_1u_x+C_2u_y$
    & 0.5476
    & The conservative diffusion placement was competitive but remained worse
      than the nonconservative placement. \\
  Eval.~63
    & $u_t=C_0\Delta u+C_1u_x+C_2u_y+C_3u_{xx}+C_4u_{yy}$
    & 0.7818
    & Independent directional diffusion corrections added freedom without
      improving transfer. \\
  Final
    & $u_t=C_0\Delta u+C_1u_x+C_2u_y$
    & 0.4944
    & Submitted the minimal archive winner after the component, nonlinearity,
      dependency, and operator-placement controls; its sealed error was 0.5114.
      \\
\end{longtable}

The trajectory contains two distinct reasoning moves. First, the Agent formed a
joint structure that no single early family supplied: diffusion plus two
independent spatial transport fields. Second, it treated the resulting leader
as a hypothesis to falsify. Removing reaction improved it, while deleting a
transport direction, coupling the directions through one field, moving the
diffusivity inside the divergence operator, and splitting the diffusion field
all reduced transfer performance.

\subsection{Compressed traces for all five discoveries}
\label{app:five-compressed-traces}

Table~\ref{tab:five-compressed-traces} summarizes each positive case through its
decisive structural transition and final selection controls.

\begin{longtable}{@{}>{\raggedright\arraybackslash}p{0.15\linewidth}>{\raggedright\arraybackslash}p{0.49\linewidth}>{\raggedright\arraybackslash}p{0.28\linewidth}@{}}
  \caption{Compressed decision traces for the five registered discoveries.
  Parenthesized pairs give the HEI evaluation index and search error.}
  \label{tab:five-compressed-traces}\\
  \toprule
  Case & Chronological structural progression & Selection evidence \\
  \midrule
  \endfirsthead
  \multicolumn{3}{l}{\footnotesize\emph{Table~\ref{tab:five-compressed-traces} continued}}\\
  \toprule
  Case & Chronological structural progression & Selection evidence \\
  \midrule
  \endhead
  \bottomrule
  \endfoot
  Groundwater flow
    & Flat heterogeneous diffusion
      $C_0\Delta h$ $(2,1.0598)$ was followed by the shared conservative
      construction $\nabla\!\cdot(C_0\nabla h)$ $(11,0.5140)$. Reaction,
      source, and heterogeneous transport extensions all increased the error.
    & A child with a unit $y$-gradient term reached 0.5132, but added
      unsupported structure. The Agent submitted the compact parent; its sealed
      error, 0.122599, was also slightly below the child's 0.122641. \\
  Variable diffusion
    & The first candidate, $C_0\Delta u$ $(1,0.2093)$, remained rank one while
      the Agent tested split directional fields (0.5045), reaction (0.3767),
      one-coordinate dependence (0.2628), and nonlinear modulation (0.2287).
    & The Agent retained the first candidate after 32 distinct scores. Its
      sealed error was 0.2890. \\
  Reaction--diffusion
    & Reaction alone $(1,10.1620)$ gave way to split diffusion plus reaction
      $(3,0.5403)$, then to a fully expanded local model $(5,0.4517)$. Linking
      the derivative and second-order terms through one shared diffusivity
      produced the conservative product-rule expansion $(7,0.4506)$.
    & Independent directional diffusivities scored 0.4592 and extra advection
      scored 0.4507. The Agent submitted the shared-field expansion; its sealed
      error was 0.4272. \\
  Advection--diffusion
    & Diffusion-only and transport-only candidates exceeded 1.0. Their joint
      construction with reaction reached 0.6588 at evaluation 31; deleting the
      reaction produced the exact three-field structure at evaluation 37 with
      error 0.4944.
    & Missing-direction, linked-drift, nonlinear-reaction, conservative-placement,
      and split-diffusion controls were all worse. The sealed error was 0.5114. \\
  Orthotropic diffusion
    & Shared isotropic diffusion $(2,0.3519)$ improved to a two-field
      nonconservative diagonal form $(4,0.2984)$. Placing each field inside its
      directional derivative produced
      $\partial_x(C_0u_x)+\partial_y(C_1u_y)$ $(15,0.2041)$.
    & The explicit shared-field product-rule expansion at evaluation 22 had the
      identical score. The Agent submitted the compact nested form; its sealed
      error was 0.2761. \\
\end{longtable}

Across the five trajectories, the same controller supports three discovery
patterns: immediate proposal followed by attempted falsification, composition
of previously incomplete operator families, and a discrete change in operator
placement or shared-field identity. HEI supplies comparable numerical evidence
for each transition, while the Agent determines which structural alternative
to construct next and which scored law to submit.

\section{Top-Five HEI Candidate Tables}
\label{app:top-five-candidates}

The following tables contain the five structurally distinct valid candidates
with the lowest search-visible cross-trajectory errors in each completed
archive. Candidates are first ordered as in
\texttt{pde history --order best}: exact cross-error, right-hand-side term
count, expression-tree size, and canonical expression. We then collapse only
exact integer no-ops, including additions multiplied by zero; the
next-ranked distinct structure fills the vacated position. Fitted coefficient
fields never enter this equivalence test.

Each table displays the original Agent-authored symbolic structure of the
representative candidate, with the DSL integer literals \texttt{N0} and
\texttt{N1} typeset as $0$ and $1$. The symbols $\checkmark$ and $\dagger$ denote operator
equivalence to the generating law and the Agent's final submission,
respectively. Structural labels were assigned after each run. Every coefficient
field is shown with its Agent-declared dependency.

{\footnotesize
\setlength{\tabcolsep}{3pt}
\renewcommand{\arraystretch}{1.18}
\begin{longtable}{@{}rr>{\raggedright\arraybackslash}p{0.49\linewidth}r>{\raggedright\arraybackslash}p{0.20\linewidth}@{}}
  \caption{Groundwater flow: five lowest-error structurally distinct valid
  candidates.}
  \label{tab:groundwater-top-five}\\
  \toprule
  Rank & Eval. & Candidate equation & $E_{\mathrm{search}}$ & Structural audit \\
  \midrule
  \endfirsthead
  \toprule
  Rank & Eval. & Candidate equation & $E_{\mathrm{search}}$ & Structural audit \\
  \midrule
  \endhead
  \bottomrule
  \endfoot
  1 & 23
    & $h_t=\partial_x(C_0(x,y)h_x)+\partial_y(C_0(x,y)h_y)
      +0h_x+1h_y$
    & 0.513184 & Integer-weighted gradient extension \\
  \textbf{2} & \textbf{11}
    & $\bm{h_t=\partial_x(C_0(x,y)h_x)+\partial_y(C_0(x,y)h_y)}$
    & \textbf{0.513982}
    & $\bm{\checkmark}$ Target; $\bm{\dagger}$ submitted \\
  3 & 14
    & $h_t=\partial_x(C_0(x,y)h_x)+\partial_y(C_0(x,y)h_y)+C_1(x,y)h$
    & 0.691988 & Extra spatial reaction \\
  4 & 15
    & $h_t=\partial_x(C_0(x,y)h_x)+\partial_y(C_0(x,y)h_y)+C_1(x,y)$
    & 0.692116 & Extra spatial source \\
  5 & 1
    & $h_t=0\Delta h$
    & 1.001298 & Agent-authored null law \\
\end{longtable}
}

{\footnotesize
\setlength{\tabcolsep}{3pt}
\renewcommand{\arraystretch}{1.18}
\begin{longtable}{@{}rr>{\raggedright\arraybackslash}p{0.49\linewidth}r>{\raggedright\arraybackslash}p{0.20\linewidth}@{}}
  \caption{Variable diffusion: five lowest-error structurally distinct valid
  candidates.}
  \label{tab:variable-diffusion-top-five}\\
  \toprule
  Rank & Eval. & Candidate equation & $E_{\mathrm{search}}$ & Structural audit \\
  \midrule
  \endfirsthead
  \toprule
  Rank & Eval. & Candidate equation & $E_{\mathrm{search}}$ & Structural audit \\
  \midrule
  \endhead
  \bottomrule
  \endfoot
  \textbf{1} & \textbf{1}
    & $\bm{u_t=C_0(x,y)\Delta u}$
    & \textbf{0.209334}
    & $\bm{\checkmark}$ Target; $\bm{\dagger}$ submitted \\
  2 & 32
    & $u_t=C_0(x,y)\Delta u(1+u^2)$
    & 0.228653 & Nonlinear modulation \\
  3 & 19
    & $u_t=C_0(x)\Delta u$
    & 0.262803 & Missing $y$ dependence \\
  4 & 21
    & $u_t=C_0(x,y)\Delta u+C_1(x,y)u_{xx}u_{yy}$
    & 0.304122 & Extra nonlinear term \\
  5 & 17
    & $u_t=C_0(x,y)\Delta u+C_1(x,y)\tanh(u)$
    & 0.365506 & Extra nonlinear reaction \\
\end{longtable}
}

{\footnotesize
\setlength{\tabcolsep}{3pt}
\renewcommand{\arraystretch}{1.18}
\begin{longtable}{@{}rr>{\raggedright\arraybackslash}p{0.49\linewidth}r>{\raggedright\arraybackslash}p{0.20\linewidth}@{}}
  \caption{Reaction--diffusion: five lowest-error structurally distinct valid
  candidates.}
  \label{tab:reaction-diffusion-top-five}\\
  \toprule
  Rank & Eval. & Candidate equation & $E_{\mathrm{search}}$ & Structural audit \\
  \midrule
  \endfirsthead
  \toprule
  Rank & Eval. & Candidate equation & $E_{\mathrm{search}}$ & Structural audit \\
  \midrule
  \endhead
  \bottomrule
  \endfoot
  \textbf{1} & \textbf{7}
    & $\begin{aligned}u_t={}&\partial_xC_0(x,y)u_x+C_0(x,y)u_{xx}\\
      &+\partial_yC_0(x,y)u_y+C_0(x,y)u_{yy}+C_1(x,y)u
      \end{aligned}$
    & \textbf{0.450614}
    & $\bm{\checkmark}$ Target; $\bm{\dagger}$ submitted \\
  2 & 18
    & $\begin{aligned}u_t={}&\partial_xC_0(x,y)u_x+C_0(x,y)u_{xx}\\
      &+\partial_yC_0(x,y)u_y+C_0(x,y)u_{yy}+C_1(x,y)u\\
      &+C_2(x,y)u_x+C_3(x,y)u_y
      \end{aligned}$
    & 0.450719 & Extra spatial transport \\
  3 & 5
    & $\begin{aligned}u_t={}&C_0(x,y)u_{xx}+C_1(x,y)u_{yy}+C_2(x,y)u_x\\
       &+C_3(x,y)u_y+C_4(x,y)u\end{aligned}$
    & 0.451745 & Five independent fields \\
  4 & 12
    & $\begin{aligned}u_t={}&\partial_xC_0(x,y)u_x+C_0(x,y)u_{xx}
       +\partial_yC_1(x,y)u_y\\&+C_1(x,y)u_{yy}+C_2(x,y)u\end{aligned}$
    & 0.459191 & Split directional diffusion \\
  5 & 16
    & $u_t=C_0(x,y)\Delta u+C_1(x,y)u_x+C_2(x,y)u_y+C_3(x,y)u$
    & 0.462034 & Independent local fields \\
\end{longtable}
}

{\footnotesize
\setlength{\tabcolsep}{3pt}
\renewcommand{\arraystretch}{1.18}
\begin{longtable}{@{}rr>{\raggedright\arraybackslash}p{0.49\linewidth}r>{\raggedright\arraybackslash}p{0.20\linewidth}@{}}
  \caption{Advection--diffusion: five lowest-error structurally distinct valid
  candidates.}
  \label{tab:advection-diffusion-top-five}\\
  \toprule
  Rank & Eval. & Candidate equation & $E_{\mathrm{search}}$ & Structural audit \\
  \midrule
  \endfirsthead
  \toprule
  Rank & Eval. & Candidate equation & $E_{\mathrm{search}}$ & Structural audit \\
  \midrule
  \endhead
  \bottomrule
  \endfoot
  \textbf{1} & \textbf{37}
    & $\bm{u_t=C_0(x,y)\Delta u+C_1(x,y)u_x+C_2(x,y)u_y}$
    & \textbf{0.494444}
    & $\bm{\checkmark}$ Target; $\bm{\dagger}$ submitted \\
  2 & 62
    & $\begin{aligned}u_t={}&\partial_x(C_0(x,y)u_x)+\partial_y(C_0(x,y)u_y)\\
      &+C_1(x,y)u_x+C_2(x,y)u_y\end{aligned}$
    & 0.547559 & Conservative diffusion placement \\
  3 & 52
    & $u_t=C_0(x,y)\Delta u+C_1(x,y)u_x+C_2(x,y)u_y+C_3(x,y)\sin(u)$
    & 0.625828 & Extra nonlinear reaction \\
  4 & 31
    & $u_t=C_0(x,y)\Delta u+C_1(x,y)u_x+C_2(x,y)u_y+C_3(x,y)u$
    & 0.658754 & Extra linear reaction \\
  5 & 39
    & $\begin{aligned}u_t={}&C_0(x,y)u_{xx}+C_1(x,y)u_{yy}+C_2(x,y)u_x\\
       &+C_3(x,y)u_y+C_4(x,y)u\end{aligned}$
    & 0.675537 & Split diffusion and reaction \\
\end{longtable}
}

{\footnotesize
\setlength{\tabcolsep}{3pt}
\renewcommand{\arraystretch}{1.18}
\begin{longtable}{@{}rr>{\raggedright\arraybackslash}p{0.49\linewidth}r>{\raggedright\arraybackslash}p{0.20\linewidth}@{}}
  \caption{Orthotropic diffusion: five lowest-error structurally distinct valid
  candidates.}
  \label{tab:orthotropic-diffusion-top-five}\\
  \toprule
  Rank & Eval. & Candidate equation & $E_{\mathrm{search}}$ & Structural audit \\
  \midrule
  \endfirsthead
  \toprule
  Rank & Eval. & Candidate equation & $E_{\mathrm{search}}$ & Structural audit \\
  \midrule
  \endhead
  \bottomrule
  \endfoot
  \textbf{1} & \textbf{15}
    & $\bm{u_t=\partial_x(C_0(x,y)u_x)+\partial_y(C_1(x,y)u_y)}$
    & \textbf{0.204067}
    & $\bm{\checkmark}$ Target; $\bm{\dagger}$ submitted \\
  2 & 22
    & $\begin{aligned}u_t={}&C_0(x,y)u_{xx}+C_1(x,y)u_{yy}\\
      &+\partial_xC_0(x,y)u_x+\partial_yC_1(x,y)u_y\end{aligned}$
    & 0.204067 & $\checkmark$ Product-rule equivalent \\
  3 & 29
    & $\begin{aligned}u_t={}&\partial_x(C_0(x,y)u_x)+\partial_y(C_1(x,y)u_y)\\
      &+C_2(x,y)u_{xx}u_{yy}\end{aligned}$
    & 0.247207 & Extra nonlinear term \\
  4 & 4
    & $u_t=C_0(x,y)u_{xx}+C_1(x,y)u_{yy}$
    & 0.298445 & Missing field derivatives \\
  5 & 2
    & $u_t=C_0(x,y)(u_{xx}+u_{yy})$
    & 0.351889 & Shared isotropic field \\
\end{longtable}
}

\section{Representative Harness Inputs}
\label{app:harness-inputs}

This section reproduces the search-visible textual inputs and controller
instructions from the registered advection--diffusion trajectory. The problem
statement uses neutral instrument language and does not name the PDE family,
coefficient fields, initial conditions, or generating law. The controller
messages are inserted into the same conversation at phase boundaries; they do
not replace the system role or the earlier dialogue.

\subsection{System prompt}
\label{app:system-prompt}

\begin{Verbatim}
You are a scientific hypothesis Agent working in an isolated Bash and
Python environment. Your goal is to infer and submit one complete governing
differential relation that explains both recorded trials.

Begin by reading PROBLEM.md, DSL.md, and data/metadata.json. Inspect the raw NPZ
arrays with Python, NumPy, and SciPy. The experimental record and measured
columns are all the domain context available: do not assume an unreported
physical identity or a known equation family. Decide which numerical
diagnostics and derivatives are informative, propose complete relations under
work/, and use the pde commands to validate and score them. Numerical scoring
is cheap; use it whenever it helps while avoiding semantic duplicates.

Treat cross-trial error as evidence, not an oracle for symbolic similarity or
uniqueness. Compare coherent structural alternatives and penalize unsupported
degrees of freedom in your scientific judgment. You may write arbitrary
analysis scripts under work/. Do not attempt to access files outside the
workspace or use the network.

Before finishing, run pde history --limit 20 --order best and then call
pde submit <candidate_id> for the relation you judge best. A prose answer is
not a submission. The evaluator never completes a partial relation.
\end{Verbatim}

\subsection{Representative \texttt{PROBLEM.md}}
\label{app:problem-prompt}

\begin{Verbatim}
# Repeated measurements from one instrumented planar specimen

A fixed rectangular array of probes records one continuous scalar response at regular measurement times. The mechanism controlling the readings is unknown.

## Recorded trials

- Trial A: Trial A records the response after preparation program A.
- Trial B: Trial B uses the same specimen and probes after preparation program B.

The `data/` directory contains the search-visible instrument records. Inspect
the raw arrays freely with Python. Arrays at the same row index form one
observation. Row order carries no physical meaning: reconstruct any tensor grid
from each row's complete coordinate tuple rather than applying a direct reshape
based only on dimension lengths. Determine a complete differential relation
that transfers across both trials. A low numerical error is evidence, not proof
of a unique governing law.
\end{Verbatim}

\subsection{Candidate-language specification}
\label{app:dsl-prompt}

\begin{Verbatim}
# Candidate language

Write one complete differential relation using the recorded column names. A
candidate is a UTF-8 text file with this shape:

```text
field-eqgpt/measured-law@1
<expression> = <expression>
```

The grammar is recursive:

```text
<expression> ::= <atom>
               | <field>
               | -<expression>
               | (<expression>)
               | <expression> <binary-op> <expression>
               | <pointwise-function>(<expression>)
               | <derivative>
<derivative> ::= D(<expression>, <coordinate-column>)
<field> ::= C<index>(<position-column>[, <position-column> ...])
<binary-op> ::= + | - | * | / | ^
```

Recorded columns are `sensor_position_1, sensor_position_2, measurement_time, sensor_reading`. An atom is a recorded column or
an integer token `N0`, `N1`, and so on. Pointwise functions are `SIN`, `COS`,
`TAN`, `COT`, `SINH`, `COSH`, `TANH`, `COTH`, `EXP`, `LOG`, `SQRT`, `ABS`,
`ARCSIN`, `ARCCOS`, `ARCTAN`, `ARCCOT`, `ARCSINH`, `ARCCOSH`, `ARCTANH`, and
`ARCCOTH`. The second argument of `D` must be a recorded position or time
column. A field's arguments must be distinct recorded position columns.

The current evaluator accepts exactly one measured time rate on the left and
no time derivative on the right. Number fields `C0`, `C1`, and so on by first
appearance without gaps. Repeating a field name means the same field and
different names mean independently fitted fields. Unknown fields must enter
linearly.

No algebraic rewrite, simplification, expansion, factorization, or term
completion is performed. Every operator remains at the syntax-tree position in
which it was written. Use `pde validate` to check syntax without spending a
score.

Commands:

```bash
pde validate work/h01.pde
pde score work/h01.pde
pde score-batch work/h01.pde work/h02.pde
pde history --limit 20 --order best
pde evidence <candidate_id>
pde submit <candidate_id>
```

`score-batch` accepts up to 32 complete candidate paths and evaluates each one
independently. The evaluator never adds missing terms. `pde evidence` writes the already-fitted coefficient fields and numerical diagnostics for one scored candidate to a JSON file under `data/evidence/`; it does not perform structural analysis.
\end{Verbatim}

\subsection{Phase-control prompts}
\label{app:controller-prompts}

The transcript below includes the observation checkpoint, the Phase-II
dual-channel instruction, the no-score reminder, and the bounded final-selection
instruction. Progress notifications that only substitute the current archive
count use the same checkpoint rule and are omitted.

\begin{Verbatim}
[Observation checkpoint]
The initial observation phase is over. Before more open-ended analysis, compare a small set of structurally distinct complete hypotheses and score at least 3 candidates. Your next Bash action must write the candidate under work/ and end exactly with `pde score work/<name>.pde`; other actions are rejected until a candidate archive contains 3 distinct scored candidates. They may be tentative: numerical feedback is cheap and is evidence rather than proof. After the minimum comparison, the run continues in a candidate-search phase using the complete conversation history; the raw observation arrays will no longer be mounted.

[Search-phase transition]
The initial data-analysis phase is complete. Continue in the same conversation using all reasoning and tool results above. The raw observation arrays are no longer available in the sandbox. Candidate files, numerical scoring, archive history, exported fitted-field evidence, and Bash/Python for analyzing that evidence remain available.

Use two complementary search operations:

- Independent construction: return to the evidence and build a complete candidate without editing or copying a scored candidate. It must represent a genuinely different structural explanation; changing only fixed numbers, signs, or a small part of a previous candidate is not independent construction. Explore the candidate language as expression trees rather than as a flat catalogue of familiar terms. Cover structural axes including but not limited to nested and non-nested compositions, grouping and factor placement across operator boundaries, additive, multiplicative, quotient, and compositional coupling, shared and independent repeated components, unknown-field identity and dependencies, operator ordering and interactions, derivative order and direction, pointwise transformations, and nonlinear combinations. This list is illustrative: identify additional structural axes from the grammar and the evidence. Do not assume candidates built from the same primitives are equivalent unless algebra or the documented evaluator semantics establishes that equivalence. Score a representative from every materially distinct, evidence-plausible family that fits the budget; do not blindly take the Cartesian product of all axes.
- Local revision: choose a scored candidate as a parent, state one evidence-based weakness or ambiguity, make one controlled change, and score the child against the parent.

Before selection, score new complete candidates from both operations. Begin this phase with independent constructions, then continue using both operations rather than allowing the current best or most recent lineage to become the template for all proposals. Scores rank only candidates actually tested; they do not rule out untested structural explanations. No particular mutation, operator, or equation family is prescribed.

[No-score reminder]
2 search turns have passed without a new scored candidate. Use the existing analysis, archive, and any exported evidence to revise a complete candidate and obtain numerical feedback. This reminder does not prescribe a mutation or equation family.

[Selection instruction]
The exploration phase is over. No new analysis or scoring is allowed. In each Bash tool call, run exactly one plain command: either `pde history --limit 20 --order best` or `pde submit <candidate_id>`. Do not use pipes, redirection, `cd`, or combine commands. If you already know your selection, submit it immediately. Lowest error is not automatically the true law; weigh physical coherence, unsupported extra terms, and simplicity. You have at most 2 selection turns.
\end{Verbatim}

\clearpage
\section{Audited Literature-Positioning Pool}
\label{app:audited-literature-positioning-pool}

\begingroup
\footnotesize
\setlength{\tabcolsep}{3pt}
\renewcommand{\arraystretch}{1.08}
\setlength{\LTleft}{0pt}
\setlength{\LTright}{0pt}
\begin{longtable}{@{}>{\raggedright\arraybackslash}p{2.30in}
  >{\raggedright\arraybackslash}p{1.10in}
  >{\raggedright\arraybackslash}p{1.55in}
  >{\raggedright\arraybackslash}p{1.25in}@{}}
\caption{Audited literature-positioning pool (87 publications).}\label{tab:positioning-literature-pool}\\
\toprule
Work / citation & Structure & Coefficient & Note \\
\midrule
\endfirsthead
\multicolumn{4}{c}{\tablename~\thetable\ (continued)}\\
\toprule
Work / citation & Structure & Coefficient & Note \\
\midrule
\endhead
\midrule
\multicolumn{4}{r}{Continued on next page}\\
\endfoot
\bottomrule
\endlastfoot
Equations of motion \citep{crutchfield_1987} & Mixed & Constant coefficients & Structure depends on representation. \\
GP structure ID \citep{gray_gp_1998} & Open-form library & Constant coefficients &  \\
GP-ODE \citep{cao_gp_2000} & Open-form library & Constant coefficients &  \\
ARE \citep{are_2007} & Open-form library & Constant coefficients &  \\
Free-form laws \citep{schmidt_lipson_2009} & Open-form library & Constant coefficients &  \\
GP dynamics \citep{gp_dynamical_systems_2014} & Open-form library & Constant coefficients &  \\
ELGP \citep{epigenetic_symbolic_dynamics_2016} & Open-form library & Constant coefficients &  \\
PDE-FIND \citep{pde_find_2017} & Closed library & Constant coefficients &  \\
SINDy \citep{sindy_2016} & Closed library & Constant coefficients &  \\
Integral sparse PDE \citep{schaeffer_pde_2017} & Closed library & Constant coefficients &  \\
PDE-Net \citep{pde_net_2018} & Closed library & Expressible by equations &  \\
PDE-Net 2.0 \citep{pde_net_2_2019} & Expandable library & Expressible by equations &  \\
DeepMoD \citep{deepmod_2021} & Closed library & Constant coefficients &  \\
EPDE \citep{epde_2019} & Expandable library & Constant coefficients &  \\
Multiphysics PDE \citep{multiphysics_pde_2019} & Closed library & Constant coefficients &  \\
SGTR \citep{sgtr_2019} & Closed library & Expressible by equations &  \\
VSI pattern formation \citep{vsi_pattern_2019} & Closed library & Constant coefficients &  \\
AI Feynman \citep{ai_feynman_2020} & Open-form library & Constant coefficients &  \\
DLGA-PDE \citep{dlga_pde_2020} & Expandable library & Constant coefficients &  \\
Homogenized equations \citep{homogenized_equations_2020} & Closed library & Constant coefficients &  \\
Low-rank grouped PDE \citep{low_rank_pde_2020} & Closed library & Mixed & Coefficient depends on representation. \\
PINN-SR \citep{pinn_sr_2021} & Closed library & Constant coefficients &  \\
Plasma reduced models \citep{plasma_reduced_models_2022} & Closed library & Constant coefficients &  \\
Symbolic GNN \citep{cranmer_symbolic_models_2020} & Open-form library & Constant coefficients &  \\
Stepwise-DLGA \citep{stepwise_dlga_2021} & Expandable library & Expressible by equations &  \\
VSI microstructure \citep{vsi_microstructure_2021} & Closed library & Constant coefficients &  \\
Weak SINDy \citep{weak_sindy_2021} & Closed library & Constant coefficients &  \\
WSINDy-PDE \citep{weak_sindy_pde_2021} & Closed library & Constant coefficients &  \\
Active-matter PDE \citep{active_matter_pde_2023} & Closed library & Constant coefficients &  \\
Bayesian VC-PDE \citep{bayesian_varcoef_2021} & Closed library & Mixed & Coefficient depends on representation. \\
HIN-PDE \citep{luo2023physics} & Closed library & Time-invariant spatial fields &  \\
PCSR \citep{pcsr_experimental_2021} & Closed library & Constant coefficients &  \\
PDE-READ \citep{pde_read_2022} & Closed library & Constant coefficients &  \\
R-DLGA \citep{r_dlga_2021} & Expandable library & Constant coefficients &  \\
SGA-PDE \citep{sga_pde_2022} & Open-form library & Constant coefficients &  \\
SINDy-AE \citep{sindy_ae_2022} & Open-form library & Constant coefficients &  \\
DHC-GEP \citep{dhc_gep_2024} & Open-form library & Constant coefficients &  \\
DISCOVER \citep{discover_2024} & Open-form library & Constant coefficients &  \\
Molecular macro-GEP \citep{molecular_macro_gep_2022} & Open-form library & Constant coefficients &  \\
Network dynamics \citep{network_dynamics_2022} & Closed library & Constant coefficients &  \\
PDE-LEARN \citep{pde_learn_2024} & Closed library & Constant coefficients &  \\
PeRCN \citep{percn_2022} & Closed library & Constant coefficients &  \\
PIC \citep{pic_pde_2023} & Closed library & Constant coefficients &  \\
SEQL / HEQL \citep{seql_heql_2023} & Open-form library & Expressible by equations &  \\
SPL \citep{spl_2023} & Open-form library & Constant coefficients &  \\
GP-IDENT \citep{gp_ident_2023} & Closed library & Time-varying / spatiotemporal fields &  \\
Granular PDE \citep{granular_pde_2023} & Closed library & Constant coefficients &  \\
Gravity-current PDE \citep{gravity_current_pde_2023} & Closed library & Constant coefficients &  \\
ODEFormer \citep{odeformer_2024} & Open-form library & Expressible by equations &  \\
R-DISCOVER \citep{r_discover_2024} & Open-form library & Constant coefficients &  \\
Rogue-wave model \citep{rogue_wave_model_2023} & Open-form library & Constant coefficients &  \\
Sparse Bayesian PDE \citep{sparse_bayesian_pde_2023} & Closed library & Constant coefficients &  \\
SPDE-VB \citep{stochastic_pde_vb_2024} & Closed library & Other irregular / stochastic fields &  \\
Unknown-equation state estimation \citep{unknown_equations_state_estimation_2023} & Closed library & Constant coefficients &  \\
Weak-PDE-LEARN \citep{weak_pde_learn_2023} & Closed library & Constant coefficients &  \\
E. coli PDE discovery \citep{ecoli_pde_discovery_2024} & Closed library & Constant coefficients &  \\
Fractional PDE \citep{fractional_pde_discovery_2025} & Expandable library & Constant coefficients &  \\
KAN-ODEs \citep{kan_odes_2024} & Mixed & Constant coefficients & Structure depends on representation. \\
LaSR \citep{lasr_2024} & Open-form library & Expressible by equations &  \\
LLM-SR \citep{llm_sr_2025} & Open-form library & Expressible by equations &  \\
LLM4ED \citep{llm4ed_2024} & Open-form library & Expressible by equations &  \\
optPDE \citep{optpde_2024} & Expandable library & Constant coefficients &  \\
PhysPDE \citep{physpde_2025} & Closed library & Constant coefficients &  \\
PINN-GP \citep{pinn_gp_2024} & Open-form library & Constant coefficients &  \\
PSE \citep{pse_2026} & Open-form library & Constant coefficients &  \\
Scientific Generative Agent \citep{scientific_generative_agent_2024} & Open-form library & Expressible by equations &  \\
SiMBA \citep{simba_2025} & Expandable library & Constant coefficients &  \\
Stochastic dynamics \citep{stochastic_dynamics_2024} & Closed library & Other irregular / stochastic fields &  \\
ABL-PDE \citep{abl_pde_2025} & Expandable library & Constant coefficients &  \\
Bayesian VC model selection \citep{bayesian_model_selection_vc_2025} & Closed library & Mixed & Coefficient depends on representation. \\
BILLIE \citep{billie_2025} & Mixed & Constant coefficients & Structure depends on representation. \\
EqGPT \citep{eqgpt_2025} & Open-form library & Expressible by equations &  \\
GraphED \citep{graphed_2025} & Open-form library & Expressible by equations &  \\
MechNNPDE \citep{mechnnpde_2025} & Expandable library & Constant coefficients &  \\
TRAFFIC-PDE-LEARN \citep{traffic_pde_learn_2025} & Closed library & Constant coefficients &  \\
WG-IDENT \citep{wg_ident_2025} & Closed library & Time-invariant spatial fields &  \\
Dynamics-aware ID \citep{dynamics_aware_pde_2026} & Closed library & Constant coefficients &  \\
Equivariant PDE discovery \citep{equivariant_pde_2026} & Closed library & Constant coefficients &  \\
Freeze-then-select \citep{freeze_then_select_2026} & Mixed & Constant coefficients & Structure depends on representation. \\
GenSR \citep{gensr_2026} & Open-form library & Constant coefficients &  \\
IGSR \citep{igsr_2026} & Open-form library & Constant coefficients &  \\
KO-PDE-IDENT \citep{ko_pde_ident_2026} & Closed library & Constant coefficients &  \\
LaSIPDE \citep{lasipde_2026} & Closed library & Constant coefficients &  \\
LLM-ODE \citep{llm_ode_2026} & Open-form library & Constant coefficients &  \\
LLM-PDESR \citep{llm_pdesr_2026} & Open-form library & Constant coefficients &  \\
LLM-PySR \citep{llm_pysr_2026} & Open-form library & Constant coefficients &  \\
Multi-source PDE \citep{joint_multisource_pde_2026} & Closed library & Constant coefficients &  \\
\end{longtable}
\endgroup

\end{document}